%% file: iclr2027_conference.tex
\documentclass{article} %
\usepackage{iclr2027_conference,times}

\input{math_commands.tex}

\usepackage{hyperref}
\usepackage{url}

\usepackage{graphicx}
\usepackage{amssymb}
\usepackage{booktabs}
\usepackage{tabularx}

\graphicspath{{figures/}{./}}

\title{Scale, Drivers, and Limits of\\Representational Alignment Across\\Large Language Models}

\author{Jialin Wu \\
Department of Cognitive Science \\
University of California, San Diego \\
La Jolla, CA 92093, USA \\
\texttt{jlwu@ucsd.edu} \\
\And
Meenakshi Khosla \\
Department of Cognitive Science \\
Department of Computer Science and Engineering \\
University of California, San Diego \\
La Jolla, CA 92093, USA \\
\texttt{mkhosla@ucsd.edu}
}

\newif\ifpreprint
\preprinttrue

\ifpreprint
  \iclrfinalcopy
  \hypersetup{pdftitle={Scale, Drivers, and Limits of Representational Alignment Across Large Language Models}, pdfauthor={Jialin Wu, Meenakshi Khosla}}
\fi
\begin{document}
\maketitle
\ifpreprint
  \lhead{Preprint. Under review.}
\fi
\begin{abstract}
Comparative analysis of large language models offers a means of identifying shared representational structure and systematic differences across them. However, aggregate measures of representational similarity, commonly used in these comparisons, can obscure how agreement varies across geometric scales and which linguistic properties support it. We introduce two complementary methods to resolve where models agree and what supports that agreement. Standard representational similarity analysis (RSA) summarizes similarity across all stimulus pairs, potentially obscuring differences in alignment between nearby and distant inputs. Neighborhood-Weighted RSA (NW-RSA) addresses this limitation by weighting stimulus pairs according to their proximity in representational space, enabling systematic comparison of alignment from local neighborhoods to global geometry on a common scale. Counterfactual RSA (CF-RSA) tests which linguistic properties support alignment by comparing intact representations in one model with representations under controlled linguistic perturbations in another. Using these tools across diverse language models, we find strong agreement in coarse representational geometry but substantial divergence in local neighborhoods. Alignment is observed both across broadly sampled natural text and within linguistic equivalence classes that preserve particular aspects of meaning or structure. Counterfactual analyses reveal selective sensitivity, with changes to logical scope reducing alignment substantially more than the tested changes to hierarchical syntax, event roles, or discourse structure. Alignment also decreases when sentence structure or lexical meaning is disrupted through word lists and jabberwocky, suggesting that the regularities of natural language support stronger convergence. These findings provide a differentiated account of representational convergence in language models and refine claims of representational universality by showing that strong overall similarity need not imply agreement in how models represent finer linguistic distinctions.
\end{abstract}

\section{Introduction}
\label{sec:intro}

A growing body of work suggests that as models scale, their internal representations converge toward a shared structure, as proposed by the \emph{Platonic representation hypothesis}~\citep{huh2024platonic}. Such convergence would mean that models trained on different data with different objectives recover a common geometry of meaning. Yet a single similarity score can be difficult to interpret because it reflects two choices. First, different measures emphasize different representational objects: Procrustes compares overall shape~\citep{williams2021generalized}, while soft-matching methods compare unit response tuning~\citep{khosla2024soft}. Models can agree in their broad layout while disagreeing about local neighborhoods. Second, the stimulus set determines which distinctions are tested. Broadly sampled sentences mostly contrast topics and concepts, whereas controlled stimuli can isolate finer semantic or syntactic differences. Agreement on natural text may therefore fail under controlled changes to meaning or form, or off the natural-language manifold. Whether representations converge depends on what is measured and what it is measured over.

\begin{figure}[t]
\centering
\includegraphics[width=0.72\linewidth]{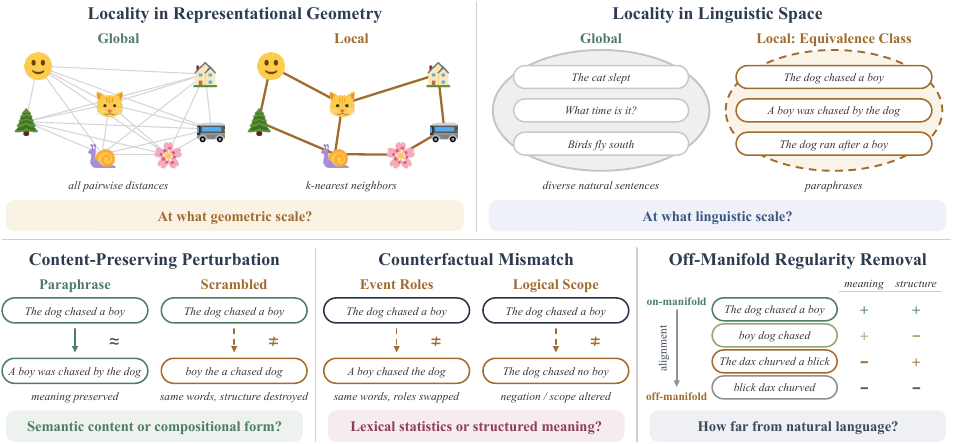}
\caption{The three axes of alignment studied here: geometric locality (NW-RSA), linguistic locality (equivalence classes), and linguistic structure and regularity (perturbations, counterfactual mismatches, and off-manifold inputs).}
\label{fig:overview}
\end{figure}

This paper distinguishes these forms of alignment through four linked questions. First, at what geometric scale do models align: the full pattern of pairwise distances or the neighborhoods around each sentence? Second, does alignment also hold within linguistic equivalence classes, such as paraphrases, sentences sharing one syntactic form, entity substitutions, discourse neighborhoods, and task-equivalent prompts? Third, which controlled linguistic perturbations disrupt alignment, revealing sensitivity to particular aspects of interpretation? Fourth, how much alignment remains when lexical meaning and sentence structure are removed? These questions separate locality in representational geometry from locality defined by linguistic properties (Figure~\ref{fig:overview}).

We address these questions through four contributions.
\textbf{(i)~A framework for comparative analysis} that examines alignment across geometric scales, within sets of sentences that share particular linguistic properties, and under controlled changes to meaning and form. It distinguishes how the comparison method and stimulus set shape conclusions about shared representations.
\textbf{(ii)~Neighborhood-Weighted RSA (NW-RSA)}, a general method for comparing representational alignment across geometric scales. By varying the weight assigned to nearby versus distant stimulus pairs, it places local and global geometric comparisons on a common scale.
\textbf{(iii)~Counterfactual RSA (CF-RSA)}, a general method for testing which stimulus properties support representational alignment. It compares one model's intact-input representational dissimilarity matrix with another's intact--perturbed dissimilarities, testing input changes relative to fixed intact references.
\textbf{(iv)~Evidence on the extent and conditions of convergence}: alignment is strong globally but weaker locally across diverse open-weight language models (Figure~\ref{fig:scale-results}). Among the tested counterfactual phenomena, logical-scope changes reduce alignment the most (Figure~\ref{fig:counterfactual}), and removing lexical meaning or sentence structure also weakens it.

These findings qualify broad interpretations of the Platonic representation hypothesis and refine claims of shared representations~\citep{klabunde2023towards,lee2025shared}.

\section{Related Work}
\label{sec:related}

\paragraph{Representational comparison.}
Similarity measures compare artificial neural networks and relate their representations to biological responses~\citep{kriegeskorte2008representational,klabunde2025similarity}. One family compares stimulus-by-stimulus relationships, including representational similarity analysis (RSA), centered kernel alignment (CKA), normalized Bures similarity, and mutual $k$-nearest-neighbor overlap~\citep{kornblith2019similarity,harvey2024duality,huh2024platonic}. These measures emphasize different aspects of representational structure. Standard RSA summarizes agreement across all pairwise dissimilarities, whereas mutual $k$-NN compares local neighbor sets. Their scores cannot be directly compared to determine whether alignment is stronger locally or globally. Topological RSA addresses a related problem by transforming distances to emphasize intermediate-scale structure~\citep{lin2024topology}. NW-RSA instead retains the original distance ranks and varies pair weights within one correlation measure, recovering global RSA when weights are uniform.

\paragraph{Representational convergence across models.}
Evidence of shared organization predates recent language-model studies. In vision, \citet{kornblith2019similarity} used CKA to compare networks across random initializations, widths, and architectures. In language, \citet{klabunde2023towards} evaluated similarity measures on commonsense reasoning and code-generation inputs, highlighting differences between models and challenges in interpreting scores. \citet{huh2024platonic} reported increasing alignment among more capable vision models and between independently trained vision and language models, motivating the Platonic representation hypothesis. After calibrating width and layer-selection effects, \citet{groger2026revisiting} found that scaling trends largely disappear for global similarity but persist for neighborhood membership. This illustrates how conclusions depend on the representational structure being compared. Naturalistic corpora introduce a further ambiguity because topic, lexical content, syntax, and meaning vary simultaneously. Our work complements these studies by combining comparisons across geometric scales with controlled linguistic inputs and perturbations.

\paragraph{Linguistic tests and counterfactual evaluation.}
Minimal pairs and contrast sets test whether individual models distinguish specific syntactic or semantic properties~\citep{marvin2018targeted,warstadt2020blimp,ribeiro2020beyond}. Probing studies examine whether those properties can be recovered from representations and how their accessibility varies across layers~\citep{hewitt2019structural,tenney2019bert}. Representational comparison raises a related question: which linguistic properties account for agreement between models? High similarity on intact inputs cannot distinguish agreement supported by lexical overlap from agreement sensitive to compositional meaning. CF-RSA adapts controlled perturbations to this question by comparing intact geometry in one model with intact--perturbed dissimilarities in another. The resulting change tests which stimulus properties support the measured agreement.

\section{Framework}
\label{sec:framework}

\begin{figure*}[t]
\centering
\raisebox{-0.5\height}{\includegraphics[width=0.52\linewidth]{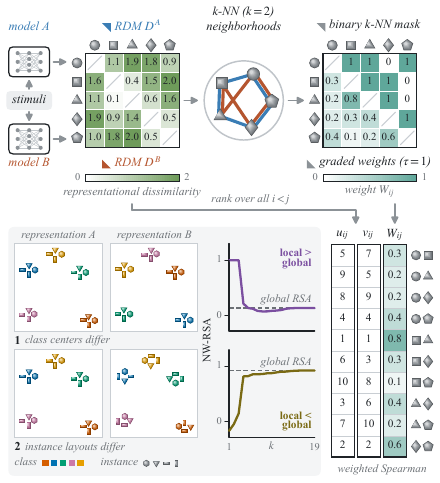}}\hfill
\raisebox{-0.5\height}{\includegraphics[width=0.44\linewidth]{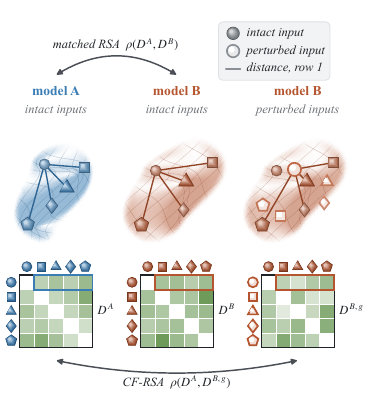}}
\caption{Two complementary RSA diagnostics (schematic; all values synthetic). Marker shapes identify the stimuli. \textbf{Left:} NW-RSA (Section~\ref{sec:nwrsa}). Each square holds two symmetric matrices: model $A$'s RDM or the binary $k$-NN mask ($k=2$) in the upper triangle, and model $B$'s RDM or the graded weights ($\tau=1$) in the lower triangle; neighborhood edges are colored by model. The shaded inset is a toy simulation with 20 points across 5 classes, where color marks the class rather than the model. In case~1, the two representations preserve the exact within-class geometry but rearrange the class centers, mimicking shared neighborhoods but a different coarse taxonomy: NW-RSA is $1.00$ for $k\le3$, while global RSA is only $0.14$. In case~2, the class centers are shared but the instances are permuted within each class, mimicking the reverse: local NW-RSA is $-0.16$ at $k=1$, while global RSA is $0.93$. \textbf{Right:} CF-RSA (Section~\ref{sec:counterfactual-rsa}). Color marks the model, and hollow markers are model $B$'s perturbed inputs $g(x_i)$, placed in its intact geometry. The schematic RDM holds the distances from each $g(x_i)$ to the intact $x_j$ (lines show row~1), excluding matched pairs; this is the transpose of $D^{B,g}$ and yields the same correlation. Correlating $D^A$ with $D^B$ gives global RSA, and correlating $D^A$ with $D^{B,g}$ gives CF-RSA.}
\label{fig:methods}
\end{figure*}

\input{framework}

\section{Experimental Setup}

\label{sec:setup}

\paragraph{Models and layers.}
The model pool contains 16 checkpoints spanning encoder-only models, GPT-2 variants, and recent Llama, Qwen, and Mistral models. This panel tests alignment across pretraining objectives, model families, and model sizes; Appendix~\ref{app:models} lists the checkpoints, and Appendix~\ref{app:assets} provides their sources. Our main analyses use the penultimate layer and mean-pool hidden states over all non-special, non-padding tokens. Appendix~\ref{app:layers} examines robustness across depth.

\paragraph{Stimuli.}
The experiment suite uses five stimulus families. The global sample contains diverse natural sentences. The equivalence-class sample contains local neighborhoods described in Section~\ref{sec:linguistic-locality}. The mismatched-perturbation sample contains controlled pairs that preserve many lexical items while changing hierarchical syntax, event roles, logical scope, or discourse structure. The perturbation sample contains generated sentences, each paired with a meaning-preserving paraphrase, a word-scrambled variant, and a word list of its content words in shuffled order. The off-manifold sample uses a $2\times2$ manipulation of lexical meaning and sentence structure. Appendix~\ref{app:stimuli} gives the sources, the generation and audit procedure, and per-experiment item counts.

\paragraph{Statistics.}
We aggregate alignment over model pairs and plot the mean with $\pm1$ SEM bands over pairs (Figures~\ref{fig:scale-results}, \ref{fig:counterfactual}). Each experiment computes its own global RSA baseline on its stimulus set and model pool, so absolute RSA values are comparable within but not necessarily across experiments.

\section{Results}
\label{sec:results}

\subsection{Geometric Alignment Depends on the Scale of Comparison}
\label{sec:result-geometry}
\begin{figure}[t]
\centering
\includegraphics[width=0.49\columnwidth,trim=0 0 0 15bp,clip]{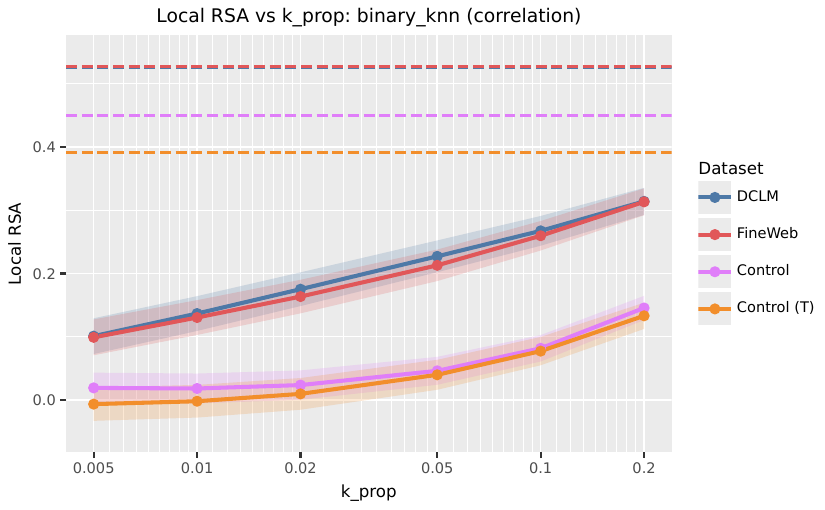}\hfill
\includegraphics[width=0.49\columnwidth,trim=0 0 0 15bp,clip]{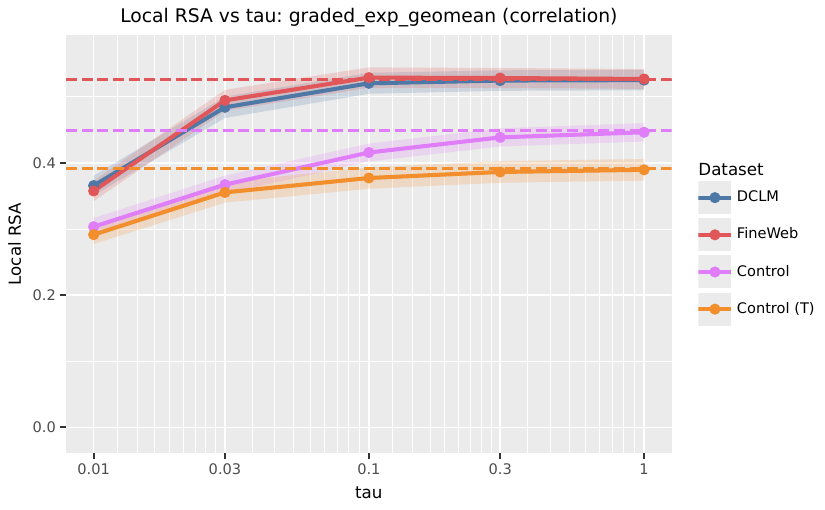}
\caption{NW-RSA across geometric scales (mean over model pairs; shaded bands $\pm1$ SEM). \textbf{Left:} binary $k$-NN neighborhoods as a proportion $k_{\mathrm{prop}}$ of stimuli. \textbf{Right:} graded temperature $\tau$. Solid lines show NW-RSA and dashed lines global RSA for DCLM, FineWeb, Control~\citep{tuckute2024driving}, and its baseline subset Control~(T).}
\label{fig:scale-results}
\end{figure}

We use NW-RSA to test whether models agree equally in local neighborhoods and global representational geometry. Figure~\ref{fig:scale-results} reports NW-RSA over model pairs across neighborhood sizes, on two naturalistic web corpora (DCLM~\citep{li2024datacomp} and FineWeb~\citep{penedo2024fineweb}), and on the Control sentence set~\citep{tuckute2024driving} as a comparison (Control~(T) is its baseline subset; Appendix~\ref{app:impl}). With $k_{\mathrm{prop}}$ denoting neighborhood size as a fraction of the stimulus set, local alignment is far below global alignment in the smallest neighborhoods: NW-RSA is near $0.10$ at $k_{\mathrm{prop}}=0.005$, against a dashed global ceiling of $\approx\!0.53$, and rises monotonically toward that ceiling as the neighborhood widens, reaching $\approx\!0.31$ at $k_{\mathrm{prop}}=0.2$. Graded weighting yields the same pattern, with NW-RSA approaching global RSA by $\tau\approx0.1$. DCLM and FineWeb show similar curves, with higher alignment than both Control sets. The same pattern holds when the two models' graded weights are combined by their maximum instead of their geometric mean (Appendix~\ref{app:metrics}).

These results show that broad geometric agreement coexists with weaker agreement among nearby inputs, measured within the same correlation objective. This differs from asking how alignment changes as models scale. \citet{groger2026revisiting} found that calibrated cross-modal neighborhood alignment increased with model capability, while global scaling trends largely disappeared. Our analysis instead compares geometric scales within model pairs: stronger global than local alignment does not imply that global alignment increases more with model capability.

\subsection{Alignment Persists Within Linguistic Equivalence Classes}
\label{sec:result-linguistic-scale}
Whereas NW-RSA defines locality by proximity in representational space, the second experiment defines it through shared linguistic properties. Within each equivalence class, we compute RSA among the neighbors of each group and among the anchors (the global baseline), holding the metric fixed while varying the linguistic scope of the inputs.

\begin{figure}[t]
\centering
\includegraphics[height=1.9in,trim=0 0 0 12.5bp,clip]{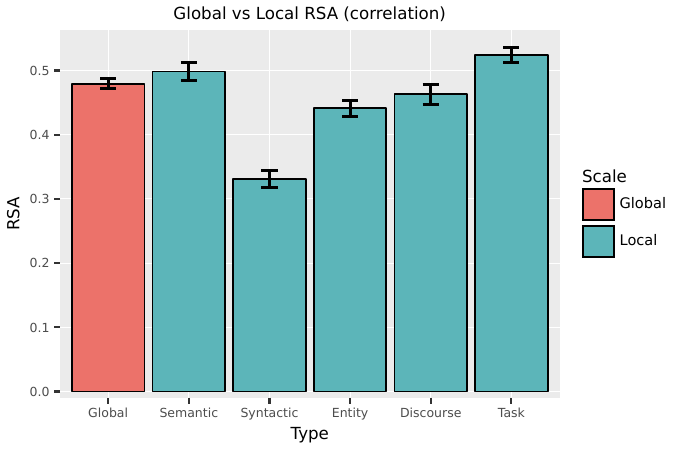}\hfill
\includegraphics[height=1.9in,trim=0 0 0 8.3bp,clip]{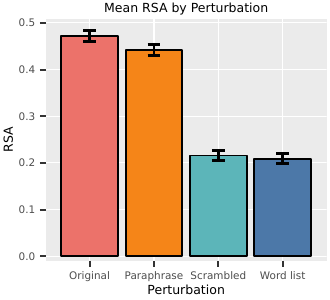}
\caption{Mean RSA over model pairs ($\pm1$ SEM). \textbf{Left:} global RSA and RSA within linguistic equivalence classes (semantic, syntactic, entity, discourse, task). \textbf{Right:} intact sentences compared with meaning-preserving paraphrases, word-scrambled inputs, and word lists.}
\label{fig:linguistic-scale}
\label{fig:perturb-paraphrase}
\end{figure}

Alignment within most of these classes is comparable to alignment across the anchor sentences (Figure~\ref{fig:linguistic-scale}, left). Relative to an RSA of $0.48$ across the anchor sentences, semantic and task classes show similar or slightly higher alignment ($0.50$ and $0.52$), discourse and entity classes show similar or slightly lower alignment ($0.46$ and $0.44$), and the syntactic class yields lower alignment ($0.33$). Thus, restricting comparisons to linguistically defined classes does not produce the pronounced reduction observed in local geometric neighborhoods. This suggests that alignment also reflects shared distinctions among inputs with similar linguistic properties. Procrustes and linear predictivity also place the syntactic class lowest and the discourse class well below the global sample (Appendix~\ref{app:metrics}). Across layers, the syntactic class is the lowest at every depth under all three metrics, while the differences among classes are largest in the first layer and narrow mainly in the early layers (Appendix~\ref{app:layers}).

\subsection{Testing Which Linguistic Properties Support Alignment with Counterfactual Analyses}
\label{sec:result-counterfactual}

The third experiment asks which linguistic distinctions support cross-model alignment. We apply Counterfactual RSA to perturbations targeting hierarchical syntax, event roles, logical scope, and discourse structure (Table~\ref{tab:mismatched_perturbations}). These perturbations alter interpretation while preserving much of the lexical content. A drop from intact RSA to CF-RSA indicates sensitivity to the tested perturbation.

\begin{table*}[t]
\caption{Examples of mismatched perturbations used for Counterfactual RSA. The paired inputs share many surface properties but differ in interpretation, isolating structure from surface form.}
\label{tab:mismatched_perturbations}
\begin{center}
\small
\setlength{\tabcolsep}{4pt}
\begin{tabularx}{\linewidth}{@{}>{\raggedright\arraybackslash}p{0.25\linewidth}XX@{}}
\toprule
\textbf{Axis} & \textbf{Model A input} & \textbf{Model B input} \\
\midrule
Hierarchical syntax~\citep{shivagunde-etal-2023-larger} &
\emph{The journalist investigated which athlete the team had recruited.} &
\emph{The journalist investigated which team the athlete had joined.} \\
Event roles~\citep{mccoy-etal-2019-right,zhang2019paws} &
\emph{The president advised the doctor.} &
\emph{The doctor advised the president.} \\
Logical scope~\citep{marelli-etal-2014-sick} &
\emph{Nobody is riding the bicycle on one wheel.} &
\emph{A person is riding the bicycle on one wheel.} \\
Discourse structure~\citep{chen-etal-2019-evaluation,mostafazadeh-etal-2016-corpus} &
\emph{The boy kicked the ball. The ball rolled away.} &
\emph{The ball rolled away. The boy kicked the ball.} \\
\bottomrule
\end{tabularx}
\end{center}
\end{table*}

\begin{figure*}[t]
\centering
\includegraphics[width=0.32\textwidth,trim=0 0 0 9.4bp,clip]{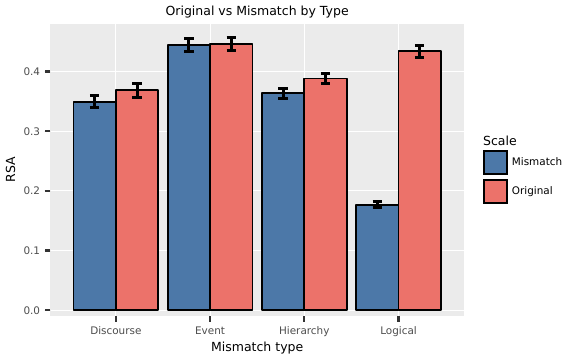}\hfill
\includegraphics[width=0.32\textwidth,trim=0 0 0 9.4bp,clip]{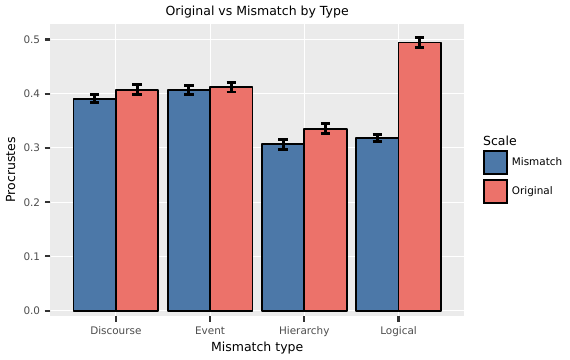}\hfill
\includegraphics[width=0.32\textwidth,trim=0 0 0 9.4bp,clip]{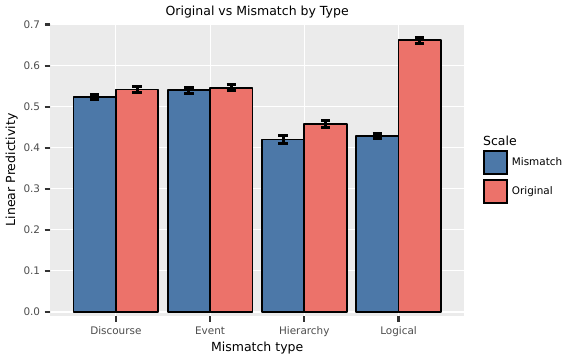}
\caption{Counterfactual (mismatch) alignment over four phenomena, pooled over model pairs ($\pm1$ SEM). Original (red) vs.\ meaning-altering mismatched (blue) pairing, under RSA, Procrustes, and linear predictivity (left to right). Logical scope collapses under mismatch in every metric. Discourse structure, event roles, and hierarchical syntax barely change.}
\label{fig:counterfactual}
\end{figure*}

We draw on established minimal-pair benchmarks. ROLE-1500~\citep{shivagunde-etal-2023-larger} reorders embedded clauses to flip who acts on whom, HANS~\citep{mccoy-etal-2019-right} and PAWS~\citep{zhang2019paws} swap subjects and objects, SICK~\citep{marelli-etal-2014-sick} introduces contradictions, and DiscoEval~\citep{chen-etal-2019-evaluation} and ROCStories~\citep{mostafazadeh-etal-2016-corpus} reverse sentence order.  We find that alignment is selectively sensitive to these changes (Figure~\ref{fig:counterfactual}). For discourse structure, event roles, and hierarchical syntax, mismatched and original pairings align almost equally well, so cross-model agreement on these inputs rests largely on shared lexical content. Logical scope is the exception: mismatching the negation and polarity structure collapses alignment (RSA $0.43\!\rightarrow\!0.18$), suggesting that here the shared geometry tracks the logical interpretation and not only the shared words. The same ordering holds for Procrustes and linear predictivity (Figure~\ref{fig:counterfactual}), with mappings fitted on intact inputs and evaluated on perturbed inputs without refitting. Across layers, logical scope shows the largest intact-versus-perturbed gap at every depth under all three metrics, but the size of this gap changes with depth. It is smallest in the first layer and grows through the middle of the network, whereas the gaps for the other three phenomena stay small throughout (Appendix~\ref{app:layers}). This growth suggests that the sensitivity to logical scope is not limited to the lexical differences between the paired inputs.  

\subsection{Testing Alignment Under Paraphrase and Word Scrambling}
\label{sec:result-perturb}

We next ask whether alignment survives changes in wording that preserve meaning and changes in word order that disrupt sentence structure. Using generated anchor sentences, we compare alignment when both models receive intact sentences with alignment when one model receives paraphrased, word-scrambled, or word-list versions (Figure~\ref{fig:perturb-paraphrase}, right; $750$ anchors). Paraphrasing produces a small reduction in RSA, from $0.47$ to $0.44$. Word scrambling produces a substantially larger reduction, to $0.22$, despite preserving the original words, and word lists yield a similar value ($0.21$). Procrustes and linear predictivity show the same ordering (Appendix~\ref{app:metrics}). These results suggest that shared representations are relatively stable under meaning-preserving rewording but sensitive to disruptions of sentence structure and compositional meaning, so lexical content alone cannot fully explain cross-model alignment. The layer-wise results show that the effect of scrambling develops with depth. In the first layer, scrambled inputs align nearly as well as intact sentences and better than paraphrases under all three metrics. With depth, paraphrases move closer to intact sentences, while scrambled inputs fall further behind them through the middle of the network (Appendix~\ref{app:layers}). Early representations thus appear to reflect which words are present more than their order.

\subsection{Alignment under Reduced Linguistic Regularity}
\label{sec:result-offmanifold}
We next ask how alignment depends on the presence of lexical meaning and sentence structure. Unlike the preceding perturbation analyses, both models receive the same inputs in each condition. A $2\times2$ design compares natural sentences~\citep{marelli-etal-2014-sick,mostafazadeh-etal-2016-corpus,zhang2019paws,cer2017semeval,shivagunde-etal-2023-larger,li2024datacomp,penedo2024fineweb} (Table~\ref{tab:off-manifold-sources}), word lists, jabberwocky sentences, and nonword lists (Table~\ref{tab:off-manifold}). Word lists retain meaningful words but remove sentence structure; jabberwocky sentences preserve sentence structure while replacing content words with pseudowords; nonword lists remove both. Three further conditions replace a single noun, verb, or adjective with a pseudoword (Appendix~\ref{app:stimuli}).

\begin{figure}[t]
\centering
\includegraphics[width=0.72\columnwidth,trim=0 0 0 13.7bp,clip]{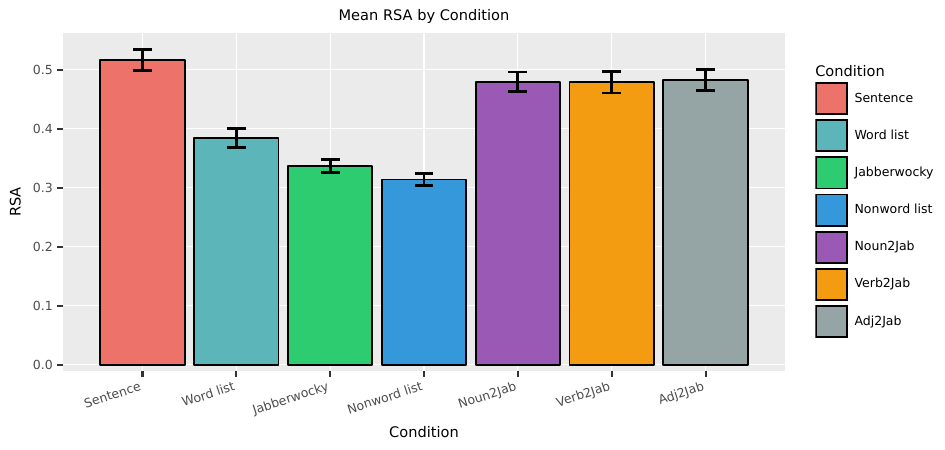}
\caption{Off-manifold alignment at the penultimate layer (mean over model pairs, $\pm1$ SEM). Noun2Jab, Verb2Jab, and Adj2Jab replace one noun, verb, or adjective with a pseudoword. Natural sentences align best, yet substantial agreement persists for nonword lists.}
\label{fig:off-manifold}
\end{figure}

Natural sentences produce the highest alignment (Figure~\ref{fig:off-manifold}; $750$ sentences per condition). Word lists and jabberwocky sentences yield lower alignment ($0.38$ and $0.34$), and nonword lists the lowest ($0.31$). Replacing a single content word has a smaller effect, and RSA remains about $0.48$ for noun, verb, and adjective substitutions. Procrustes and linear predictivity show the same pattern, and the human-authored stimuli of \citet{regev2024neural} show a similar reduction (Appendices~\ref{app:stimuli} and~\ref{app:metrics}). Across layers, natural sentences in our stimuli align best at every depth, and their advantage over word lists, jabberwocky, and nonword lists is larger in later layers than in the first (Appendix~\ref{app:layers}). On the human-authored stimuli, sentences overtake nonword lists only after the earliest layers.

These results show that alignment depends on linguistic regularity, yet substantial alignment persists even for nonword lists, so shared geometry extends beyond meaningful, well-formed language.

\section{Discussion}

\label{sec:discussion}

Our findings show that representational convergence across language models depends on the geometric scale of comparison and the linguistic properties of the inputs. Models share broad representational geometry but agree substantially less on relative dissimilarities within local neighborhoods. This qualifies broad interpretations of the Platonic representation hypothesis~\citep{huh2024platonic}, since strong overall similarity need not extend to finer representational distinctions.

Alignment also persists within linguistic equivalence classes, where particular aspects of meaning or structure are held fixed, so shared geometry does not arise solely from broad differences among heterogeneous sentences. Because linguistic equivalence does not require geometric proximity, this result does not contradict the local--global NW-RSA gap; locality defined by representational distance differs from locality defined by linguistic properties.

The perturbation analyses show that alignment is selectively sensitive to linguistic interpretation. Logical-scope changes reduce it far more than the tested changes to syntax, event roles, or discourse structure, and it survives paraphrasing more readily than word scrambling, suggesting sensitivity to sentence structure and compositional meaning beyond lexical content. Alignment is also highest for natural sentences and decreases when lexical meaning or sentence structure is removed, yet substantial agreement persists for nonword lists, so shared geometry cannot be equated with shared linguistic interpretation.

Across layers, the scrambling effect is small in the first layer and develops through the middle layers, whereas logical-scope and regularity effects are present from the first layer and grow later, suggesting that these sensitivities are not confined to initial lexical representations.

Claims of convergence should therefore specify the geometric scale, stimulus distinctions, and input properties to which alignment is sensitive. NW-RSA and Counterfactual RSA provide tools for testing these conditions across neural systems, moving beyond whether models align to what they share and under which conditions.

\subsection*{AI use statement}

We used DeepSeek-V4-Flash to generate the equivalence-class and paraphrase stimuli, DeepSeek-V4-Pro to audit them (Appendix~\ref{app:stimuli}), and Llama-Guard-4-12B for content-safety screening (Appendix~\ref{app:safety}). AI assistants supported coding, literature searches, and editing for grammar, phrasing, and formatting. They did not contribute research ideas or experimental design. We reviewed all AI-assisted work. Generated stimuli were checked by code and audited by a separate LLM, and the equivalence-class data were also spot-checked by hand (Appendix~\ref{app:stimuli}). The authors take full responsibility for the content of this paper.

\subsection*{Ethics statement}

This work analyzes internal representations of existing language models using textual stimuli. It involves no human subjects and trains no new models. We use two kinds of stimuli and treat their risks separately. For \textbf{external public datasets} (STS-B~\citep{cer2017semeval}, PAWS~\citep{zhang2019paws}, the counterfactual benchmarks, and the web corpora), we use them within their licenses (Appendix~\ref{app:assets}) and additionally screen datasets with Llama~Guard~\citep{inan2023llama,metallamaguard4} and with rule-based filters to rule out offensive content and personally identifiable information (PII), so their reuse raises no new ethical concern beyond what their original releases already address. For the \textbf{stimuli we generate} (the equivalence-class neighborhoods and the paraphrase pairs), the generation and audit prompts explicitly forbid offensive, sexual, political, brand-, and PII-related content, with flagged items dropped or rewritten; the off-manifold conditions are derived by rule from human-written public sentences and contain no generated text. The full procedure is detailed in Appendix~\ref{app:safety}. Two residual risks remain. First, automated filtering can miss cases, so a small amount of offensive content or PII may persist despite screening. Second, deliberately controlling the stimulus distribution, by narrowing it to specific equivalence classes, perturbations, and off-manifold conditions, can reduce how well our findings generalize to naturally occurring text. We therefore keep our claims tied to the evaluated models, layers, and datasets. Appendix~\ref{app:limitations} discusses further limitations.

\subsection*{Reproducibility statement}

Section~\ref{sec:framework} defines the alignment metrics, and Section~\ref{sec:setup} describes the experimental setup. Appendix~\ref{app:models} describes the model panel. Appendix~\ref{app:stimuli} documents stimulus sources, generation and audit procedures, rejection statistics, and rule-based off-manifold transformations; Appendix~\ref{app:safety} describes content-safety filtering. Appendix~\ref{app:impl} provides representation-extraction procedures, metric parameters, aggregation rules, and the compute and software environment. Appendix~\ref{app:assets} lists the external assets and their licenses. These sections provide the methodological details supporting reproduction of the reported analyses. Upon acceptance, we will publicly release our code and the datasets constructed in this study, the latter subject to the applicable licenses of their source datasets.

\bibliographystyle{iclr2027_conference}
\bibliography{iclr2027_conference}

\appendix
\clearpage
\suppressfloats[t]

\section{Model Panel}
\label{app:models}

The main comparisons span three model families. \textbf{Encoder-only masked language models} (BERT-base, RoBERTa-base, MiniLM-L12, ALBERT-base, BERT-large, RoBERTa-large, DeBERTaV3, and ELECTRA) test whether the alignment patterns hold for bidirectional encoders trained with masked-token objectives or related discriminative pretraining. \textbf{Decoder-only language models} (GPT-2 and GPT-2-medium) connect the analysis to left-to-right language modeling without modern instruction tuning or very large scale. \textbf{Recent open-weight LLMs} (Llama-3.2-1B, Llama-3.2-3B, Qwen-2.5-0.5B, Qwen-2.5-3B, Llama-2-7B, and Mistral-7B) test whether the same patterns persist for larger decoder-only models from recent open families. Exact Hugging Face identifiers, parameter counts, and hidden sizes will be released with our code. Unless noted otherwise, we read hidden states from each model's penultimate layer and mean-pool over all non-special, non-padding tokens.
\section{Stimulus Construction}
\label{app:stimuli}

Table~\ref{tab:dataset} summarizes the source and scale of the stimuli behind each experiment. The counterfactual pairs are selected from existing datasets along the four axes of Table~\ref{tab:mismatched_perturbations} (hierarchical syntax, event roles, logical scope, and discourse structure). The stimulus families that we construct ourselves are described below. The equivalence-class neighborhoods and the paraphrase pairs are generated and audited by separate LLMs, and the off-manifold conditions are derived by rule from human-written sentences; Figure~\ref{fig:data-pipeline} summarizes the first and the last. Content-safety screening is described in Appendix~\ref{app:safety}.

\begin{table}[t]
\caption{Stimulus source and scale for each experiment. The generated equivalence-class and paraphrase datasets and the off-manifold conditions will be released. The four counterfactual datasets and the web corpora are public (Table~\ref{tab:assets}).}
\label{tab:dataset}
\begin{center}
\footnotesize
\setlength{\tabcolsep}{3pt}
\renewcommand{\arraystretch}{1.2}
\begin{tabularx}{\linewidth}{@{}XXX@{}}
\toprule
Experiment & Source & Scale \\
\midrule
Geometric scale (\S\ref{sec:result-geometry}) & DCLM~\citep{li2024datacomp}, FineWeb~\citep{penedo2024fineweb}, Control~\citep{tuckute2024driving} & $10{,}000$ sents each, Control $2000$ (Control\,(T) $1000$) \\
Linguistic scale (\S\ref{sec:result-linguistic-scale}) & Generated equivalence classes & $5$ classes $\times\,150$ groups (anchor $+\,50$ neighbors) \\
Counterfactual (\S\ref{sec:result-counterfactual}) & ROLE, HANS+PAWS, SICK, DiscoEval+ROCStories & $750$ pairs per phenomenon \\
Perturbation (\S\ref{sec:result-perturb}) & Generated paraphrase pairs & $750$ anchors $\times$ \{paraphrase, scrambled, word list\} \\
Off-manifold (\S\ref{sec:result-offmanifold}) & Derived from sentences & $750$ sents $\times\,7$ conditions \\
\bottomrule
\end{tabularx}
\end{center}
\end{table}

\begin{figure*}[t]
\centering
\includegraphics[width=\linewidth]{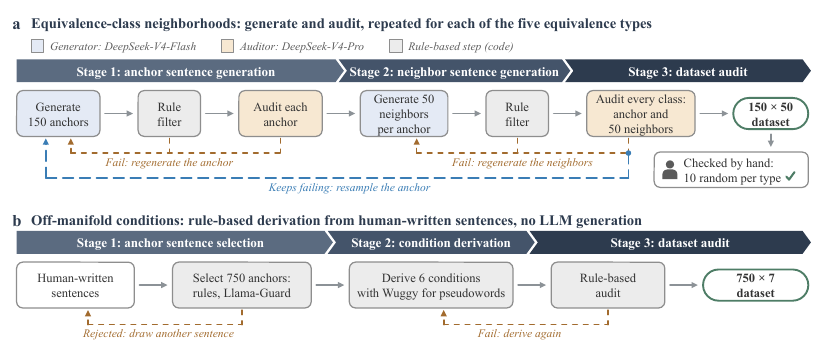}
\caption{Construction of the stimuli we build ourselves. \textbf{(a)} Equivalence-class neighborhoods, built separately for each of the five equivalence types in three stages. (1) The generator (DeepSeek-V4-Flash) proposes 150 anchors, code checks their word count, and the auditor (DeepSeek-V4-Pro) checks each anchor; rejected anchors are regenerated, for up to five audit rounds. (2) The generator writes 50 neighbors for each anchor, and code checks the length, duplicates, and wording diversity of every set. (3) The auditor reviews the complete dataset, reading each anchor together with its 50 neighbors; the neighbors are audited at this stage. A group rejected by either check receives a new neighbor set (brown dashed paths); after four failed sets or five failed audit rounds, its anchor is resampled (blue path), at most three times. Finally, we randomly chose 10 examples per type and found that all of them satisfied the constraints. \textbf{(b)} Off-manifold conditions, built in three stages. (1) Candidate sentences from seven public datasets are normalized, filtered, and screened, and a rejected candidate is replaced from the pool. (2) Six conditions are derived from each of the 750 anchors by rule, with pseudowords from Wuggy and no LLM generation. (3) A rule-based audit re-derives every condition and checks it against its anchor; a condition that fails is derived again. The released dataset holds the 750 anchors in seven conditions. The source datasets are listed in Table~\ref{tab:off-manifold-sources}, and the filter criteria and derivation rules are given in the text.}
\label{fig:data-pipeline}
\end{figure*}

\paragraph{Equivalence-class neighborhoods.}
To increase robustness and reduce potential stylistic patterns, we built the equivalence-class datasets using a multi-LLM pipeline, with diversity requirements in the prompts. For each of the five equivalence types evaluated in Section~\ref{sec:result-linguistic-scale} (semantic, syntactic, entity, discourse, and task equivalence), there are 150 anchor sentences, each with exactly 50 neighboring sentences. We used DeepSeek-V4-Flash for the generator and DeepSeek-V4-Pro for the auditor~\citep{deepseekai2026deepseekv4}. Every prompt states the property that the type holds fixed and the dimensions that may vary (Section~\ref{sec:linguistic-locality}), together with the content restrictions of Appendix~\ref{app:safety}. Dataset construction proceeded in three stages (Figure~\ref{fig:data-pipeline}a).

First, the generator produced candidate anchors, which were then audited by another LLM. Before the audit, code rejected anchors that were not in English or fell outside the word-count range of the type. The auditor checked grammaticality, construct validity, cross-group distinctness, and other requirements: each anchor had to be a single grammatical sentence expressing a concrete, unambiguous proposition that can support many equivalent neighbors, with no near-duplicates among the 150 anchors. Failed anchors were regenerated with the auditor's feedback and audited again, for up to five rounds. Second, the generator produced exactly 50 neighbors separately for each anchor, and code checked every neighbor set for word counts, duplicates, repeats of the anchor, and wording diversity. For the semantic, discourse, and task types, a set needed at least 12 distinct three-word openings and no opening shared by more than 8 neighbors; for the syntactic type, which holds grammatical form fixed, no content word of the anchor could reappear in a neighbor. A set that failed was regenerated with feedback, and if it still failed after four attempts, the anchor was resampled. Third, after all 150 groups had been assembled, the complete dataset was sent to the auditor, so the neighbors were audited at this stage rather than group by group. This dataset-level audit screened for violations of the intended equivalence relation, unnatural outputs, duplicate wording, repetitive templates, and insufficient cross-group distinctness. Groups that failed had their neighbors regenerated with the audit feedback, while accepted groups were preserved as context. If a group repeatedly failed, its anchor was resampled, and the group was reconstructed. An anchor was resampled at most three times, whether the trigger was a neighbor set that failed four times or a group that failed five audit rounds. The audit-repair cycle continued until the target groups passed. At the end, for each type, we randomly chose 10 examples and found that all of them satisfied the constraints of their type.

\paragraph{Rejection statistics.}
Table~\ref{tab:rejections} counts the rejections recorded in the generation logs. Anchor columns count single sentences, and neighbor columns count whole sets of 50 neighbors. The last column is the number of discarded sentences recorded in the logs, that is, the anchor rejections plus 50 times the rejected neighbor sets. Invalid calls are neighbor-generation calls that returned malformed or incomplete output. The logs do not record how many sentences these calls contained, so they are excluded from the last column, which is therefore a lower bound. The syntactic type required by far the most regeneration, and the discourse type the least.

\begin{table}[t]
\caption{Rejections during equivalence-class construction. \emph{Rule} and \emph{LLM} count candidates rejected by the code checks and by the auditor, and \emph{resampled} counts anchors replaced after their group kept failing. Anchor columns count sentences, and neighbor columns count sets of 50 neighbors. \emph{Invalid calls} returned malformed or incomplete neighbor sets and are not counted in the last column, which gives the discarded sentences recorded in the logs and is a lower bound.}
\label{tab:rejections}
\begin{center}
\small
\setlength{\tabcolsep}{3pt}
\renewcommand{\arraystretch}{1.15}
\begin{tabular}{@{}lrrrrrrr@{}}
\toprule
& \multicolumn{3}{c}{Anchors} & \multicolumn{2}{c}{Neighbor sets} & Invalid & Discarded \\
\cmidrule(lr){2-4} \cmidrule(lr){5-6}
Type & rule & LLM & resampled & rule & LLM & calls & sentences \\
\midrule
Task & 0 & 74 & 1 & 36 & 32 & 10 & 3,475 \\
Semantic & 0 & 78 & 4 & 85 & 90 & 29 & 8,832 \\
Syntactic & 92 & 67 & 39 & 229 & 88 & 45 & 16,048 \\
Entity & 43 & 105 & 0 & 8 & 17 & 22 & 1,398 \\
Discourse & 0 & 27 & 0 & 0 & 0 & 3 & 27 \\
\midrule
Total & 135 & 351 & 44 & 358 & 227 & 109 & 29,780 \\
\bottomrule
\end{tabular}
\end{center}
\end{table}

\paragraph{Paraphrase pairs.}
The perturbation analysis (Section~\ref{sec:result-perturb}) uses a separate dataset of 750 anchor sentences, each paired with one paraphrase that preserves its propositional meaning and truth conditions while varying syntax, lexical choice, style, and information packaging. The same generator and auditor as above produced the pairs. The generator wrote them in batches of 25, each batch assigned one of 20 everyday topic domains (e.g., household routines, travel, science, sports), and code checked that every sentence was English, 18--35 words long, and not a duplicate of any other sentence in the dataset after normalization. The auditor then checked all anchors together for grammaticality, concreteness, and distinctness, and checked each pair, in batches of 25, for changes to who did what to whom, quantity, negation, modality, time, location, cause, or outcome. Failed pairs were regenerated with the audit feedback and audited again, for up to three repair rounds. The two remaining conditions are derived from each anchor by rule: the word-scrambled variant keeps all of its word tokens in random order with punctuation removed, and the word list keeps only its content words and negation tokens in random order.

\paragraph{Off-manifold conditions.}
For the off-manifold stimuli, we built a dataset of 750 anchor sentences. To reduce stylistic patterns, we used human-written sentences from established datasets as the anchor sentences (Table~\ref{tab:off-manifold-sources}; Figure~\ref{fig:data-pipeline}b). Candidate anchors were normalized, filtered, and screened: a candidate had to be a single complete clause and could not duplicate another candidate, sentences from the web and STS-B sources that carry proper nouns were cross-checked with named-entity recognition, and sentences from all sources were screened with Llama-Guard-4-12B~\citep{inan2023llama,metallamaguard4}. The generation of the other conditions does not involve LLMs. For the jabberwocky condition, eligible content words were replaced with pseudowords generated by Wuggy~\citep{keuleers2010wuggy}, which matches the length and the character- and phoneme-level statistics of the word it replaces, while function words, punctuation, and word order were kept. When Wuggy found no candidate for a word under our strict settings, Wuggy was rerun with looser neighborhood and bigram tolerances, and an anchor whose words stayed unresolvable was dropped and replaced from the candidate pool. Every occurrence of a word maps to the same pseudoword throughout the release. The word-list condition shuffled the content words of the original sentences, while nonword lists came from shuffling pseudowords; both dropped function words. Finally, the noun-, verb-, and adjective-jabberwocky conditions replaced exactly one eligible noun (including proper nouns), lexical verb, or adjective, respectively, with its pseudoword from the jabberwocky sentence, while leaving the remainder of the sentence unchanged; the word was selected with spaCy part-of-speech tags (Appendix~\ref{app:impl}). A rule-based audit then re-derived every condition from the anchors and checked that the conditions stay aligned item by item, that a word list and a nonword list contain exactly the content words or the pseudowords of their source sentence, that each graded condition replaces exactly one target of the intended part of speech, and that the pseudoword assignment is consistent; a condition that failed the audit was derived again. Table~\ref{tab:off-manifold} summarizes which conditions keep meaning and structure.

\begin{table}[t]
\caption{Human-written sources of the 750 off-manifold anchor sentences. Licenses and citations for each source are listed in Table~\ref{tab:assets}.}
\label{tab:off-manifold-sources}
\begin{center}
\footnotesize
\begin{tabular}{@{}llr@{}}
\toprule
Source & Subset & Sentences \\
\midrule
SICK & contradiction subset & 242 \\
ROCStories & training set & 213 \\
PAWS & label-0 subset & 132 \\
STS-B & filtered & 69 \\
ROLE-1500 & deduplicated & 41 \\
DCLM & web corpus & 30 \\
FineWeb & web corpus & 23 \\
\midrule
Total & & 750 \\
\bottomrule
\end{tabular}
\end{center}
\end{table}

\begin{table}[t]
\caption{Off-manifold stimulus families. The $2\times2$ base removes word-level meaning and sentence structure in separate steps. Three graded conditions replace one content word with a pseudoword while retaining structure and most meaning.}
\label{tab:off-manifold}
\begin{center}
\small
\begin{tabular}{p{3.0cm}cc}
\toprule
\textbf{Condition} & \textbf{Meaning} & \textbf{Structure} \\
\midrule
Sentences & Yes & Yes \\
Word lists & Yes & No \\
Jabberwocky & No & Yes \\
Nonword lists & No & No \\
\midrule
Noun$\rightarrow$pseudoword & Partial & Yes \\
Verb$\rightarrow$pseudoword & Partial & Yes \\
Adj$\rightarrow$pseudoword & Partial & Yes \\
\bottomrule
\end{tabular}
\end{center}
\end{table}

\begin{figure}[t]
\centering
\includegraphics[width=0.32\linewidth,trim=0 0 0 10.8bp,clip]{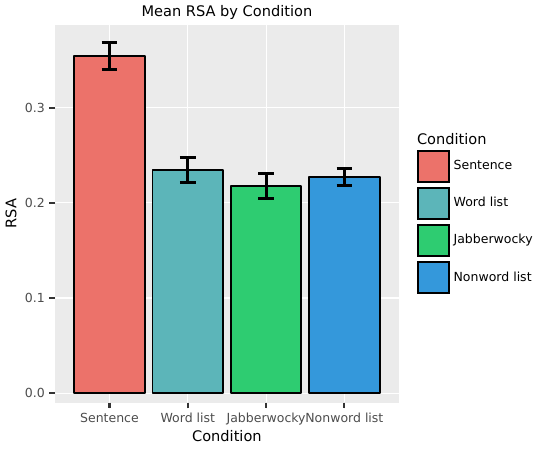}\hfill
\includegraphics[width=0.32\linewidth,trim=0 0 0 10.8bp,clip]{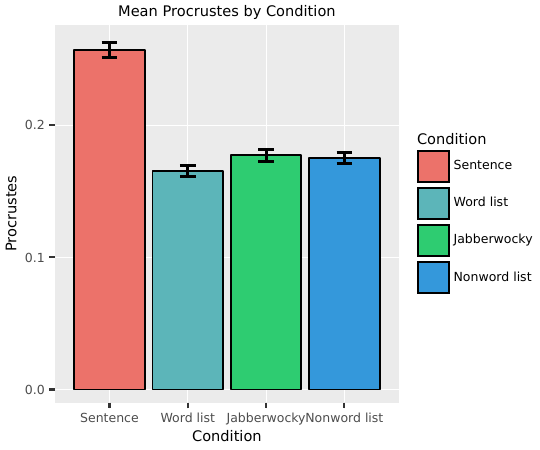}\hfill
\includegraphics[width=0.32\linewidth,trim=0 0 0 10.8bp,clip]{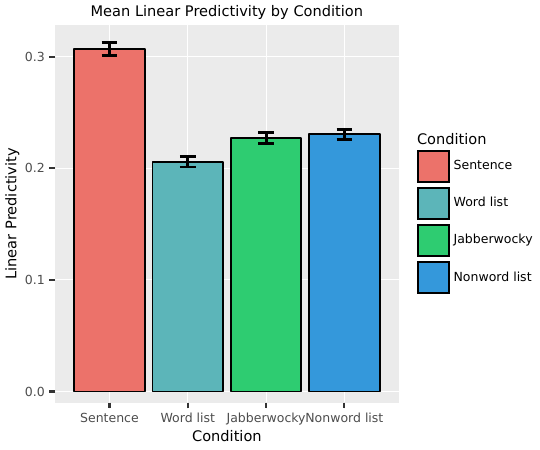}
\caption{Off-manifold alignment on the human-authored stimuli of \citet{regev2024neural} (mean over model pairs, $\pm1$ SEM), read at the penultimate layer. \textbf{Left:} RSA; \textbf{middle:} Procrustes; \textbf{right:} linear predictivity.}
\label{fig:swjn}
\end{figure}

\paragraph{Check on human-authored stimuli.}
We also tested the human-authored SWJN dataset of \citet{regev2024neural}, whose word and nonword lists mix items across sentences. Under all three metrics (Figure~\ref{fig:swjn}), the differences among word lists, jabberwocky sentences, and nonword lists are smaller than on our stimuli (Figure~\ref{fig:off-manifold}). Layer-wise results are reported in Appendix~\ref{app:layers}.

\section{Data and Content Safety}
\label{app:safety}

\paragraph{Sources.} The stimuli come from three sources. The global sample, the counterfactual pairs, and the off-manifold anchor sentences draw on public datasets (Table~\ref{tab:assets}). The equivalence-class neighborhoods and the paraphrase pairs are generated with an instruction-tuned LLM and audited by a second LLM against the constraints below. The off-manifold conditions are derived by rule from the human-written off-manifold anchor sentences (Appendix~\ref{app:stimuli}).

\paragraph{Content filtering.} To keep the datasets we will release free of offensive content and personal information, public-dataset text is screened at extraction time with Llama-Guard-4-12B~\citep{inan2023llama,metallamaguard4} and samples flagged unsafe are dropped, with named-entity recognition (spaCy) flagging person mentions for review. The generation and audit prompts for the equivalence-class and paraphrase stimuli explicitly forbid offensive, sexual, political, brand-, and PII-related content, and a separate audit pass removes violations. The off-manifold conditions are derived by rule from the human-written anchor sentences and add no generated text.

\clearpage
\suppressfloats[t]
\section{Alternative Alignment Measures}
\label{app:metrics}

The main text reports RSA for most experiments. This appendix reports a variant of the graded NW-RSA weights and the Procrustes and linear-predictivity versions of the equivalence-class, perturbation, and off-manifold analyses, all read at the penultimate layer. The Procrustes and linear-predictivity versions of the counterfactual analysis appear in Figure~\ref{fig:counterfactual}, and those of the human-authored off-manifold stimuli in Figure~\ref{fig:swjn}.

\begin{figure}[t]
\centering
\includegraphics[width=0.6\linewidth,trim=0 0 0 15bp,clip]{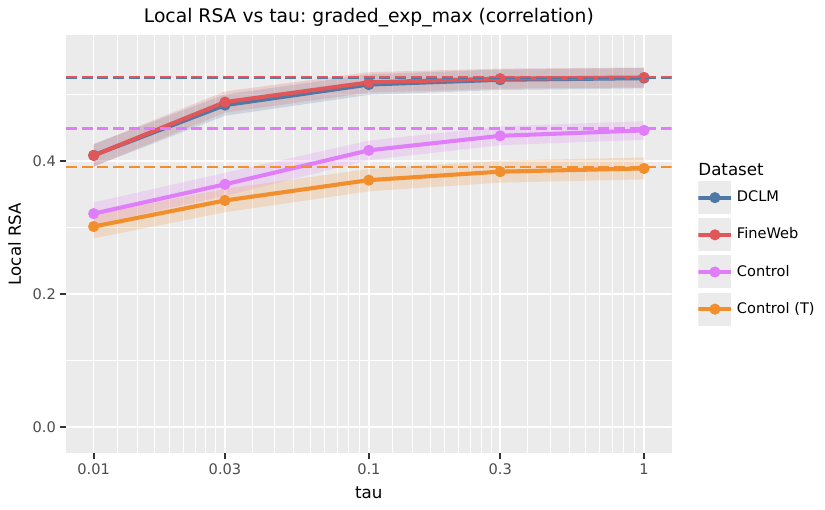}
\caption{Graded NW-RSA with the two models' weights combined by their maximum instead of their geometric mean (mean over model pairs; shaded bands $\pm1$ SEM). Dashed lines mark global RSA.}
\label{fig:scale-variants}
\end{figure}

\paragraph{Geometric scale.}
Figure~\ref{fig:scale-variants} repeats the graded NW-RSA sweep of Figure~\ref{fig:scale-results} but combines the two models' weights by their maximum, $W_{ij}=\max\big(\exp(-D^A_{ij}/\tau),\exp(-D^B_{ij}/\tau)\big)$, instead of their geometric mean. The curves keep the same shape: on the web corpora, NW-RSA starts higher at the smallest $\tau$ but stays well below global RSA there, and it approaches global RSA at a similar $\tau$.

\begin{figure}[t]
\centering
\includegraphics[width=0.49\linewidth,trim=0 0 0 12.5bp,clip]{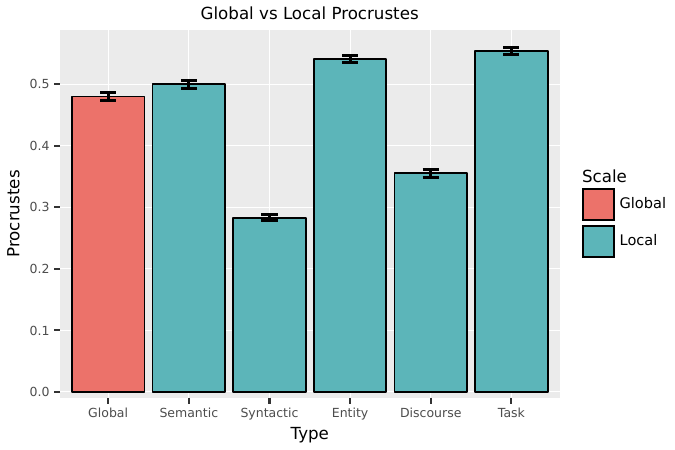}\hfill
\includegraphics[width=0.49\linewidth,trim=0 0 0 12.5bp,clip]{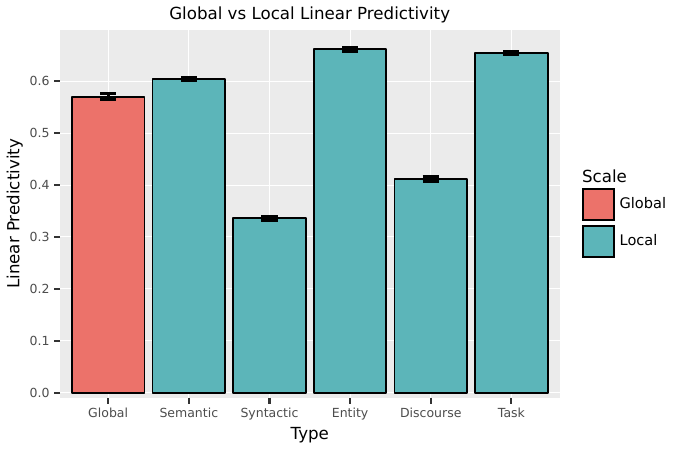}
\caption{Linguistic-scale analysis under mapping-based metrics (mean over model pairs, $\pm1$ SEM). \textbf{Left:} Procrustes; \textbf{right:} linear predictivity.}
\label{fig:linguistic-metrics}
\end{figure}

\paragraph{Linguistic equivalence classes.}
Under Procrustes and linear predictivity (Figure~\ref{fig:linguistic-metrics}), the semantic, entity, and task classes align at or above the global sample.

\begin{figure}[t]
\centering
\includegraphics[width=0.35\linewidth,trim=0 0 0 8.3bp,clip]{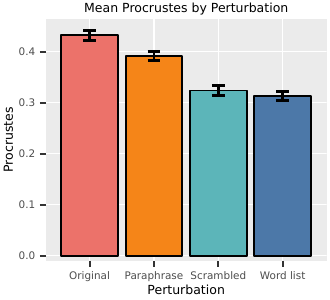}\hspace{0.08\linewidth}
\includegraphics[width=0.35\linewidth,trim=0 0 0 8.3bp,clip]{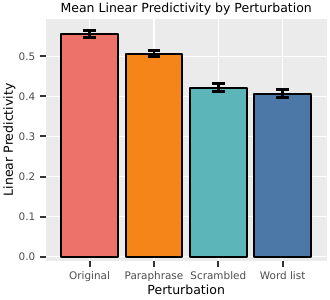}
\caption{Paraphrase and word-scrambling analysis under mapping-based metrics (mean over model pairs, $\pm1$ SEM). \textbf{Left:} Procrustes; \textbf{right:} linear predictivity.}
\label{fig:perturb-metrics}
\end{figure}

\paragraph{Paraphrase and word scrambling.}
Under Procrustes and linear predictivity (Figure~\ref{fig:perturb-metrics}), the drops for scrambled inputs and word lists are smaller than under RSA.

\begin{figure}[t]
\centering
\includegraphics[width=0.75\linewidth,trim=0 0 0 13.7bp,clip]{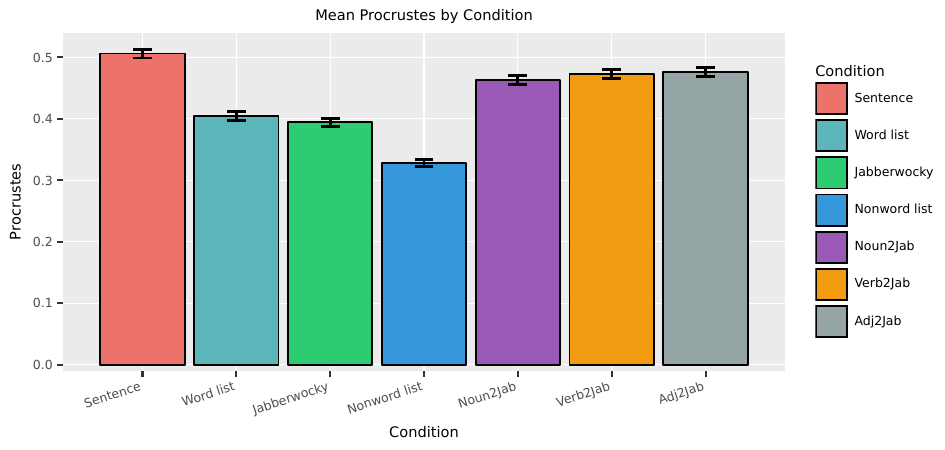}\\[4pt]
\includegraphics[width=0.75\linewidth,trim=0 0 0 13.7bp,clip]{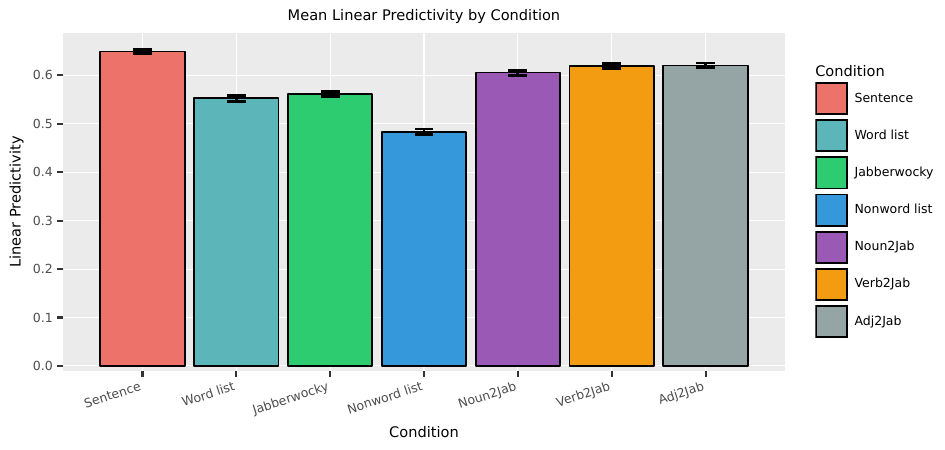}
\caption{Off-manifold alignment under mapping-based metrics (mean over model pairs, $\pm1$ SEM). \textbf{Top:} Procrustes; \textbf{bottom:} linear predictivity.}
\label{fig:offmanifold-metrics}
\end{figure}

\paragraph{Reduced linguistic regularity.}
Figure~\ref{fig:offmanifold-metrics} gives the Procrustes and linear-predictivity versions of Figure~\ref{fig:off-manifold}.

\clearpage
\section{Alignment Across Layers}
\label{app:layers}
\suppressfloats[t]

The main analyses read representations at the penultimate layer. Here we ask whether the results change with depth, repeating the equivalence-class, counterfactual, perturbation, and off-manifold analyses across relative depth under RSA, Procrustes, and linear predictivity. Relative depth is counted in each model's effective functional units (e.g., transformer blocks). Relative depth $1$ is the last such unit, which is model-dependent and differs by a normalization step from the penultimate hidden state read for the bar comparisons, so values at depth $1$ need not match the bar plots. Lines show the mean over model pairs with $\pm1$ SEM bands.

\begin{figure}[t]
\centering
\includegraphics[width=0.49\linewidth]{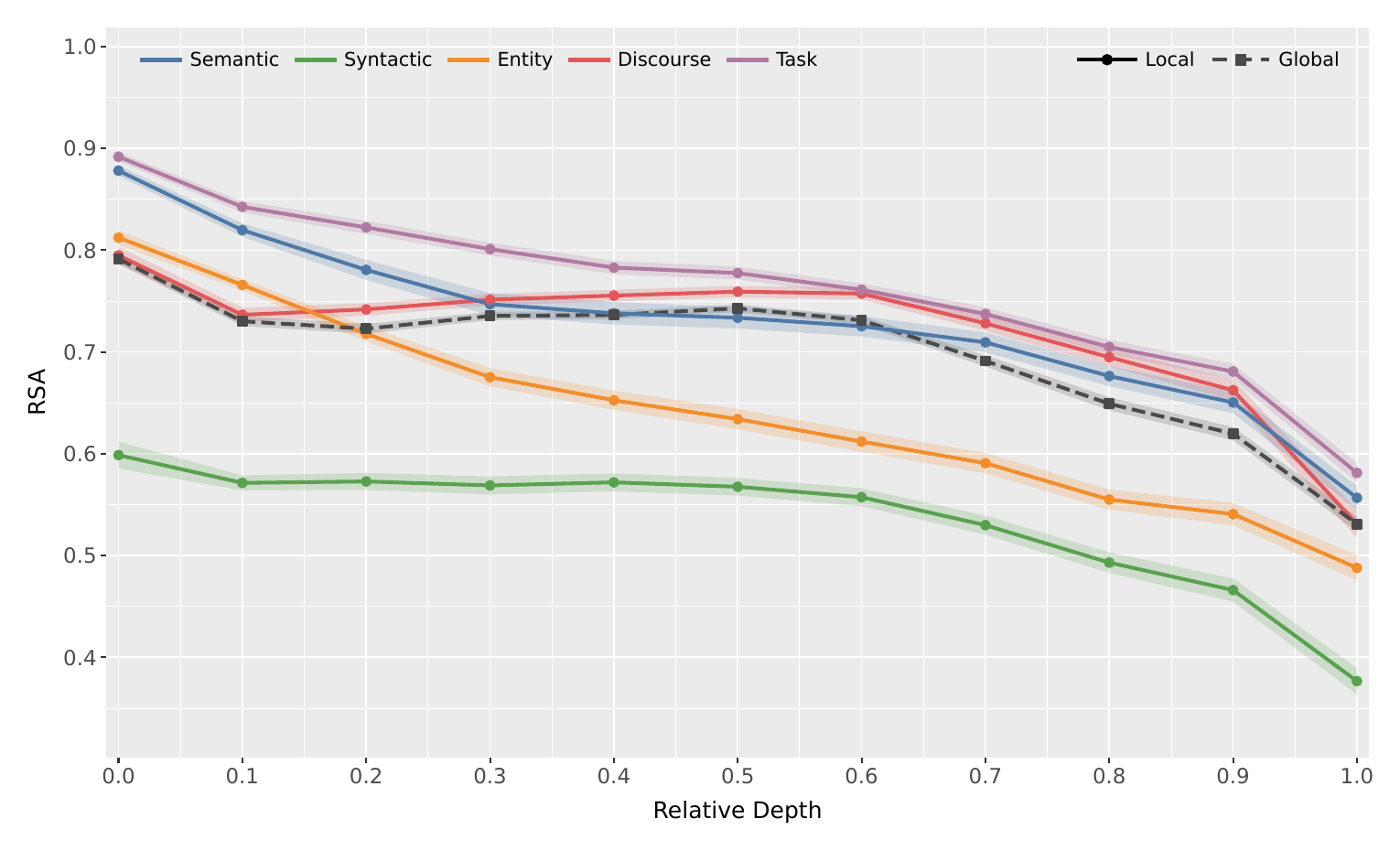}\hfill
\includegraphics[width=0.49\linewidth]{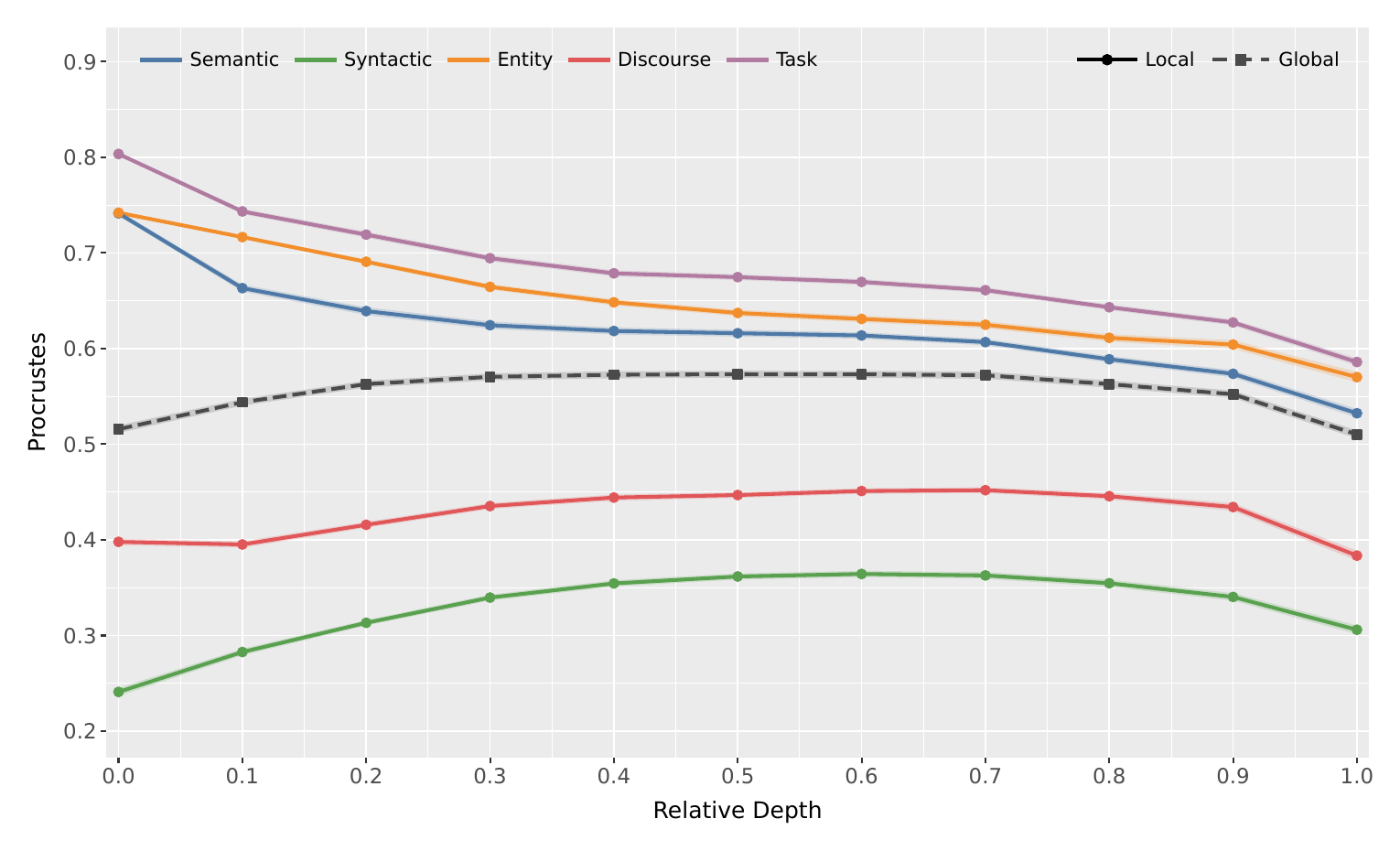}\\[4pt]
\includegraphics[width=0.49\linewidth]{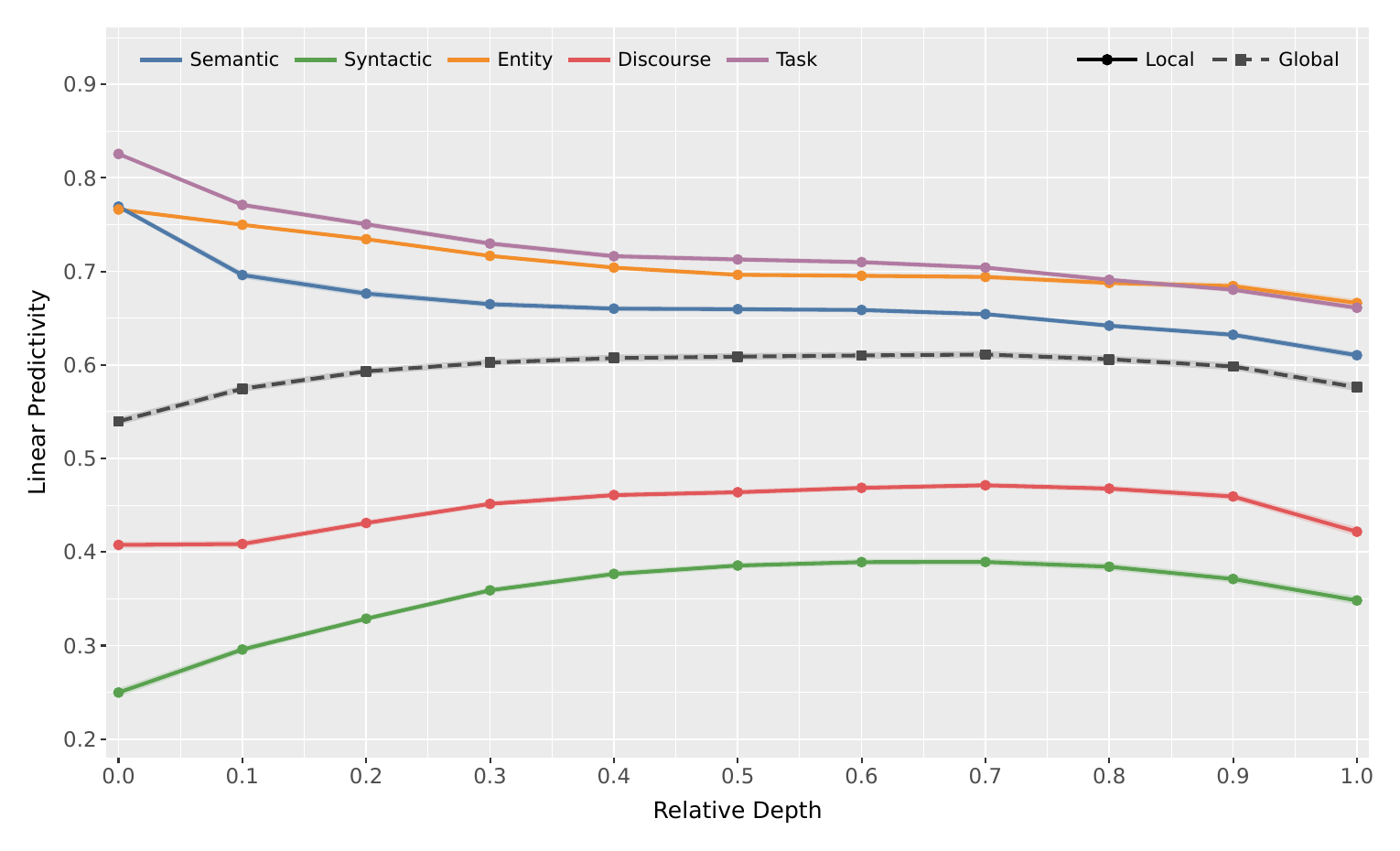}
\caption{Layer-wise linguistic-scale analysis by relative network depth. \textbf{Top left:} RSA; \textbf{top right:} Procrustes; \textbf{bottom:} linear predictivity. Solid lines are the equivalence classes and the dashed line is the global sample. The syntactic class is the lowest at every depth, and the spread across classes narrows with depth.}
\label{fig:linguistic-layerwise}
\end{figure}

\paragraph{Linguistic equivalence classes.}
The metrics differ in how the spread across classes narrows with depth (Figure~\ref{fig:linguistic-layerwise}). Under RSA, alignment declines overall with depth for every class and for the global sample, and the entity class drops from above the global sample to below it after the earliest layers. Under Procrustes and linear predictivity, the semantic, entity, and task classes start high and decline, while the syntactic class, and to a lesser extent the discourse class and the global sample, rise through the middle layers.

\begin{figure}[t]
\centering
\includegraphics[width=0.49\linewidth]{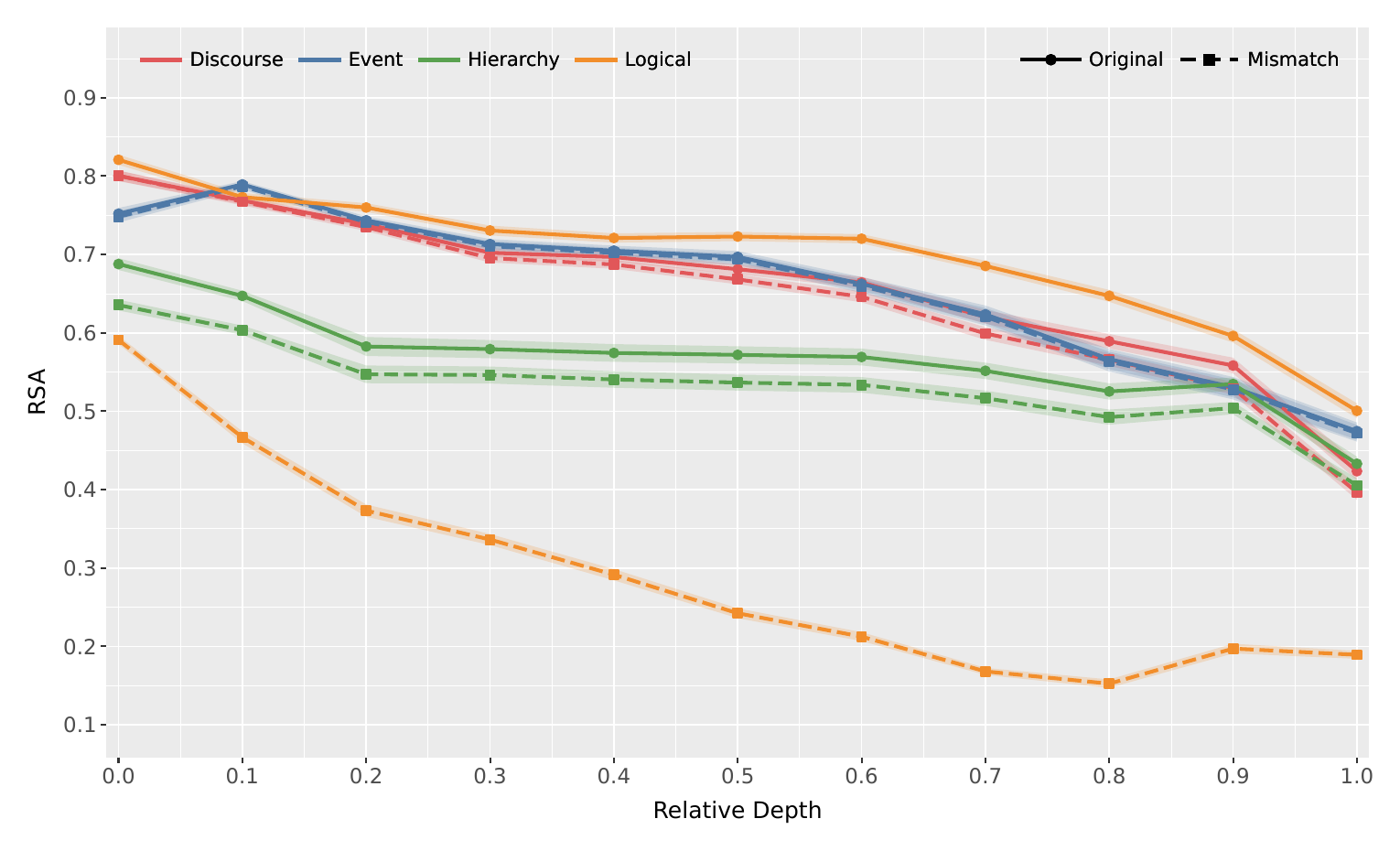}\hfill
\includegraphics[width=0.49\linewidth]{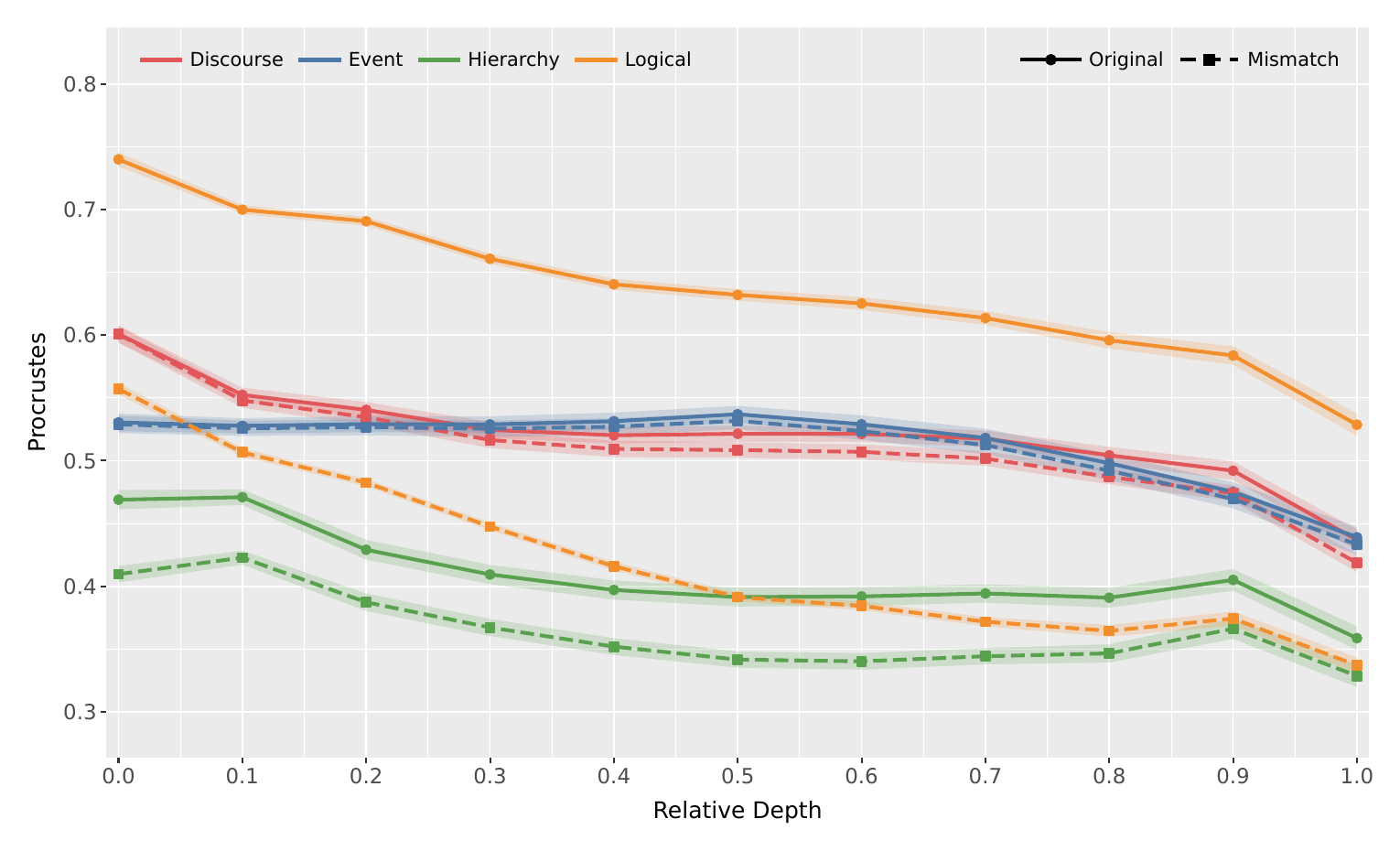}\\[4pt]
\includegraphics[width=0.49\linewidth]{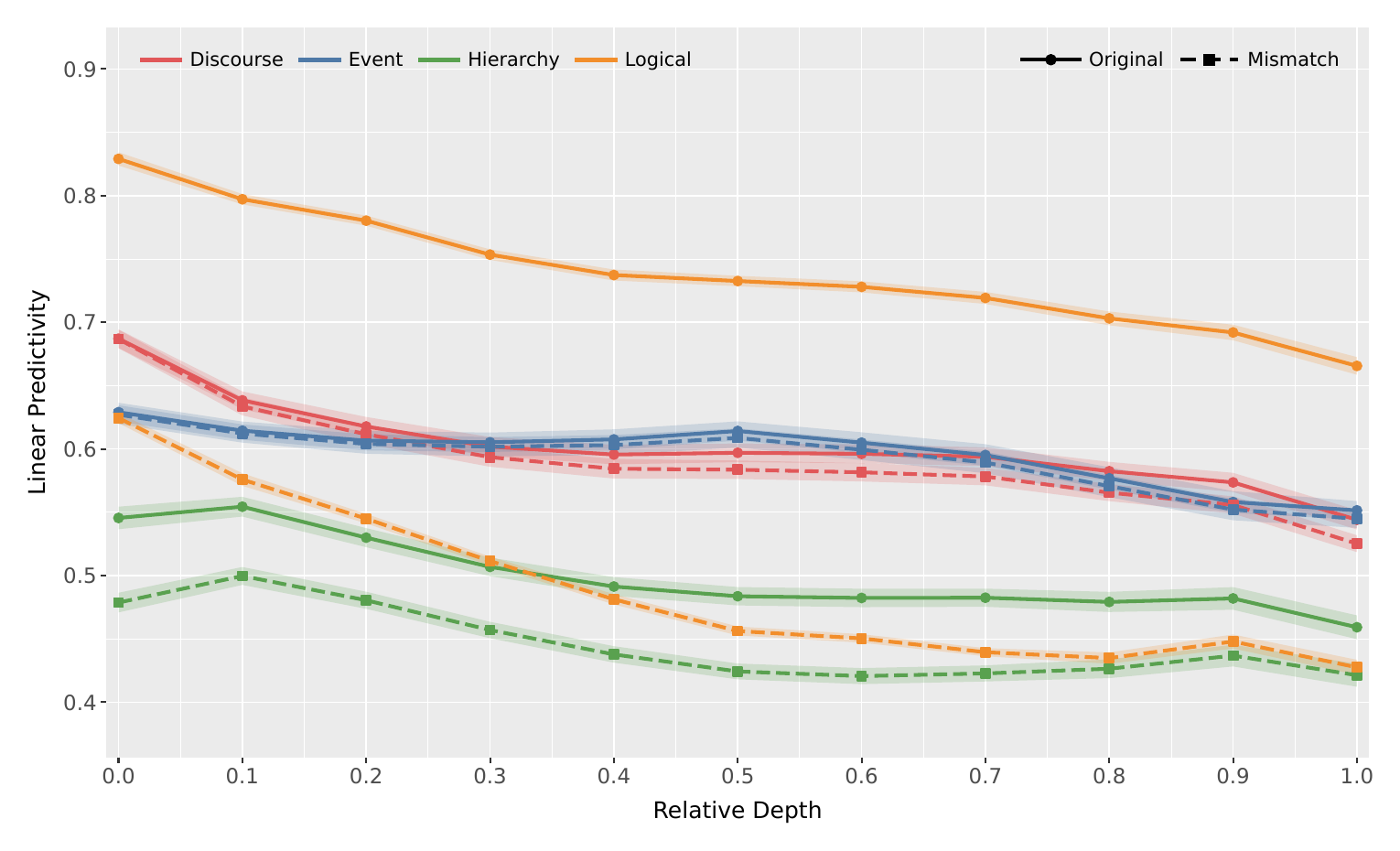}
\caption{Layer-wise counterfactual contrast by relative network depth. \textbf{Top left:} RSA; \textbf{top right:} Procrustes; \textbf{bottom:} linear predictivity. Solid lines are the original pairing, dashed the mismatched pairing. The logical-scope gap is the largest at every depth and widens through the middle layers, while the discourse-structure, event-role, and hierarchical-syntax gaps remain small.}
\label{fig:mismatch-layerwise}
\end{figure}

\paragraph{Counterfactual perturbations.}
After growing through the middle of the network, the logical-scope gap narrows again toward the output, and its growth is largest under RSA (Figure~\ref{fig:mismatch-layerwise}). Among the other three phenomena, the event-role gap is negligible throughout, and the discourse-structure gap is absent at the first layer and grows only slightly with depth.

\begin{figure}[t]
\centering
\includegraphics[width=0.49\linewidth]{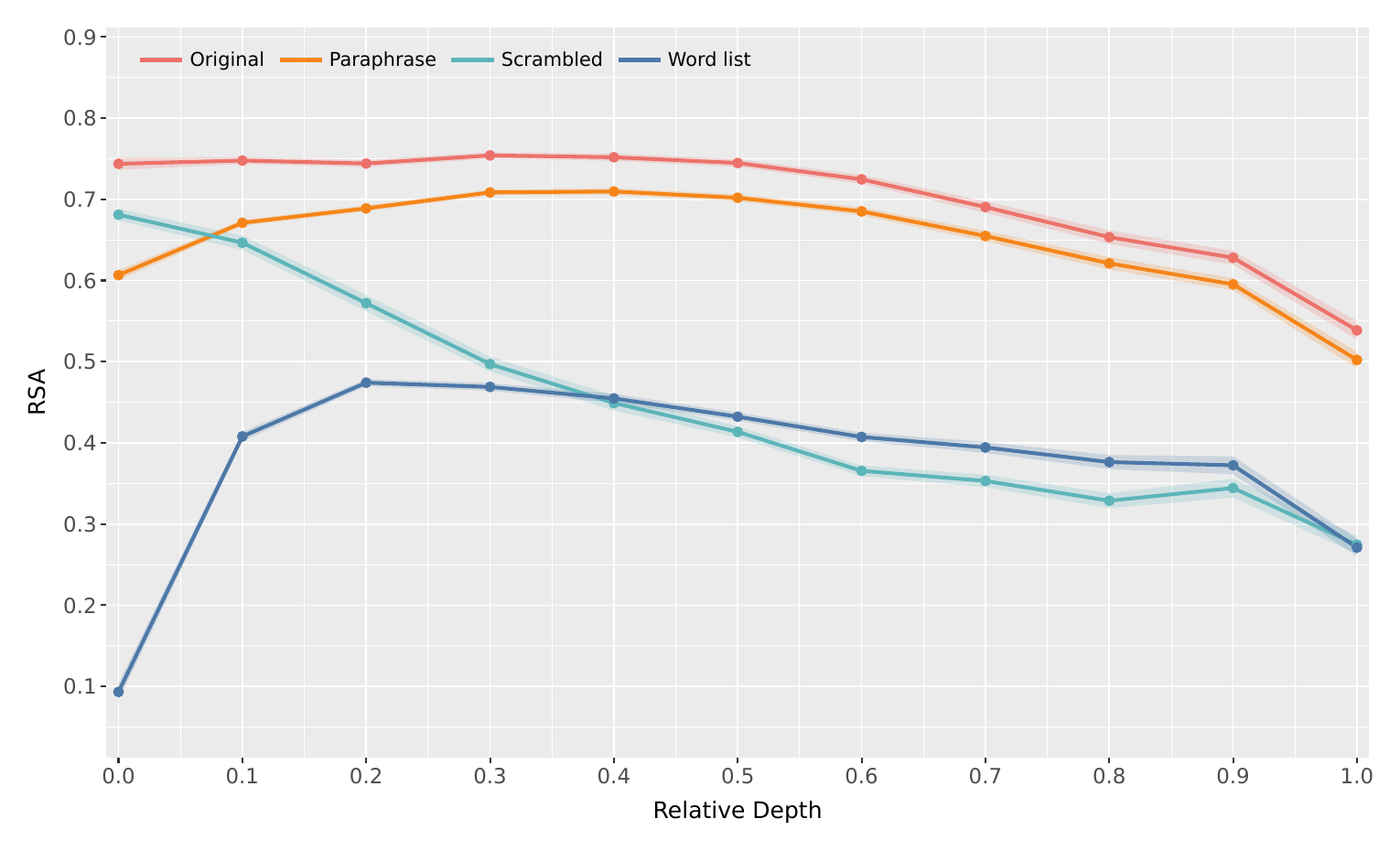}\hfill
\includegraphics[width=0.49\linewidth]{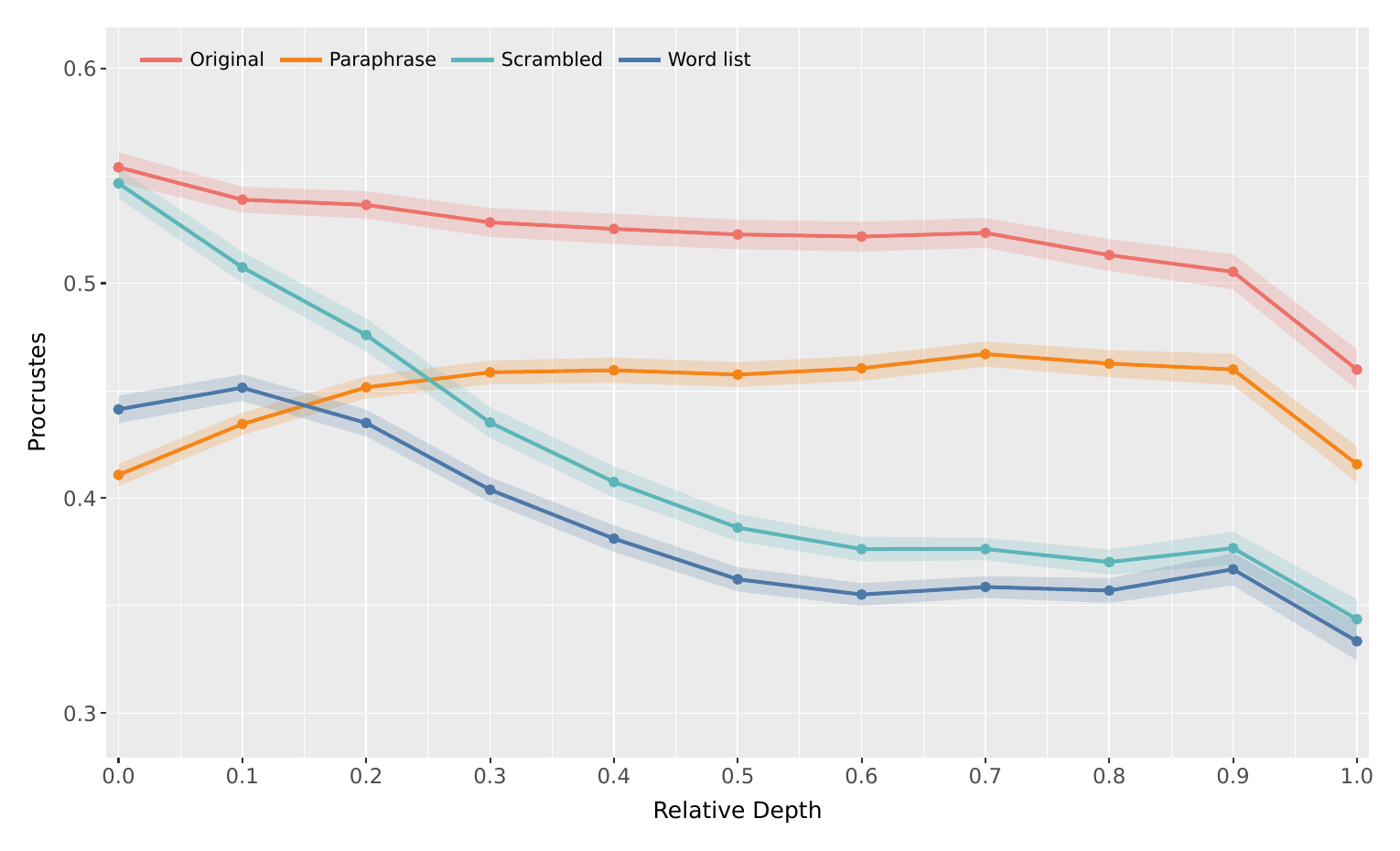}\\[4pt]
\includegraphics[width=0.49\linewidth]{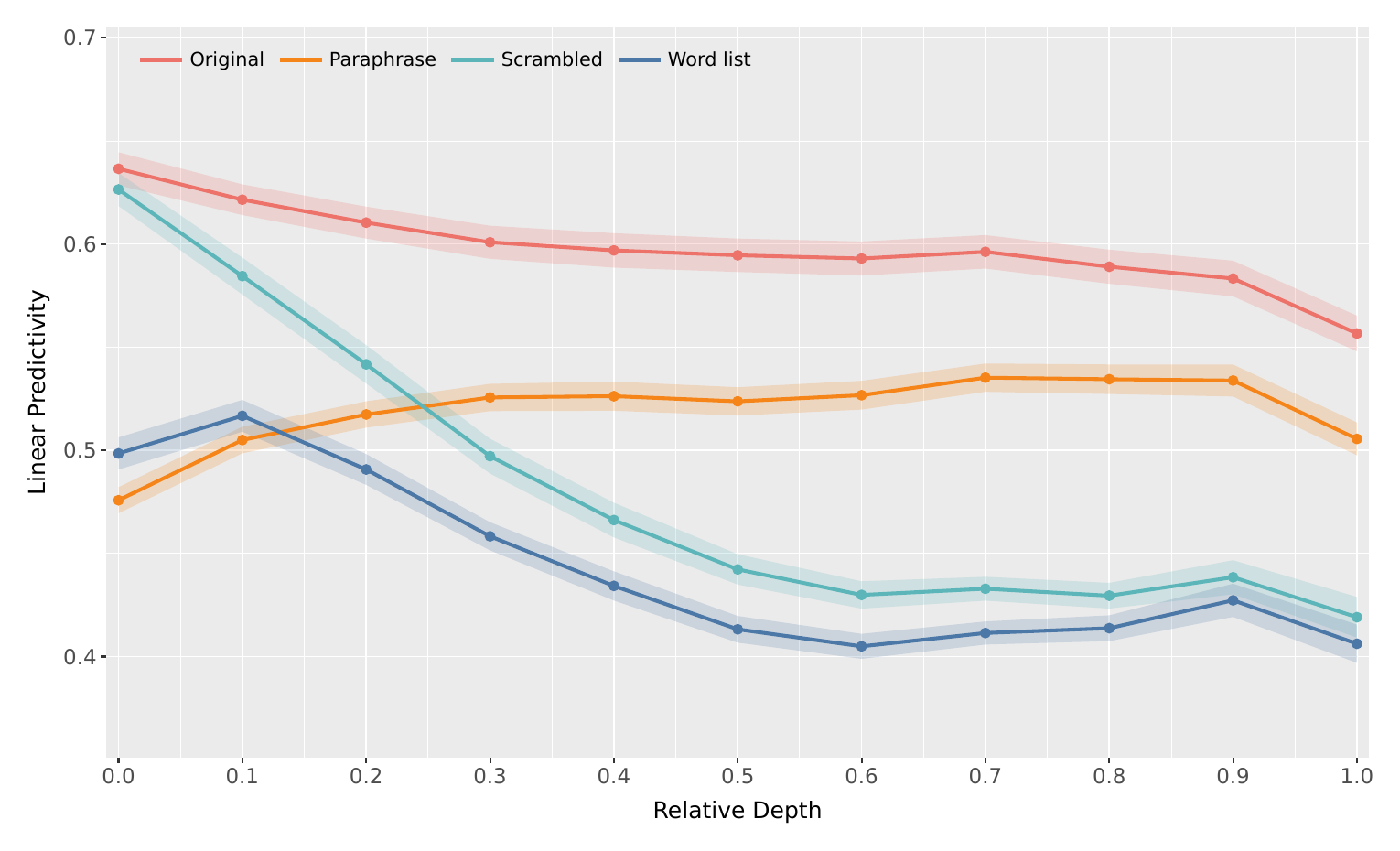}
\caption{Layer-wise paraphrase and word-scrambling analysis by relative network depth. \textbf{Top left:} RSA; \textbf{top right:} Procrustes; \textbf{bottom:} linear predictivity. Scrambled inputs align nearly as well as intact sentences at the first layer and decline with depth, while paraphrases move closer to intact sentences.}
\label{fig:perturb-layerwise}
\end{figure}

\paragraph{Paraphrase and word scrambling.}
Paraphrases overtake scrambled inputs within the early layers (Figure~\ref{fig:perturb-layerwise}). Beyond the earliest layers, word lists stay close to scrambled inputs; under RSA, they align far below the other conditions in the earliest layers, especially the first.

\begin{figure}[t]
\centering
\includegraphics[width=0.49\linewidth]{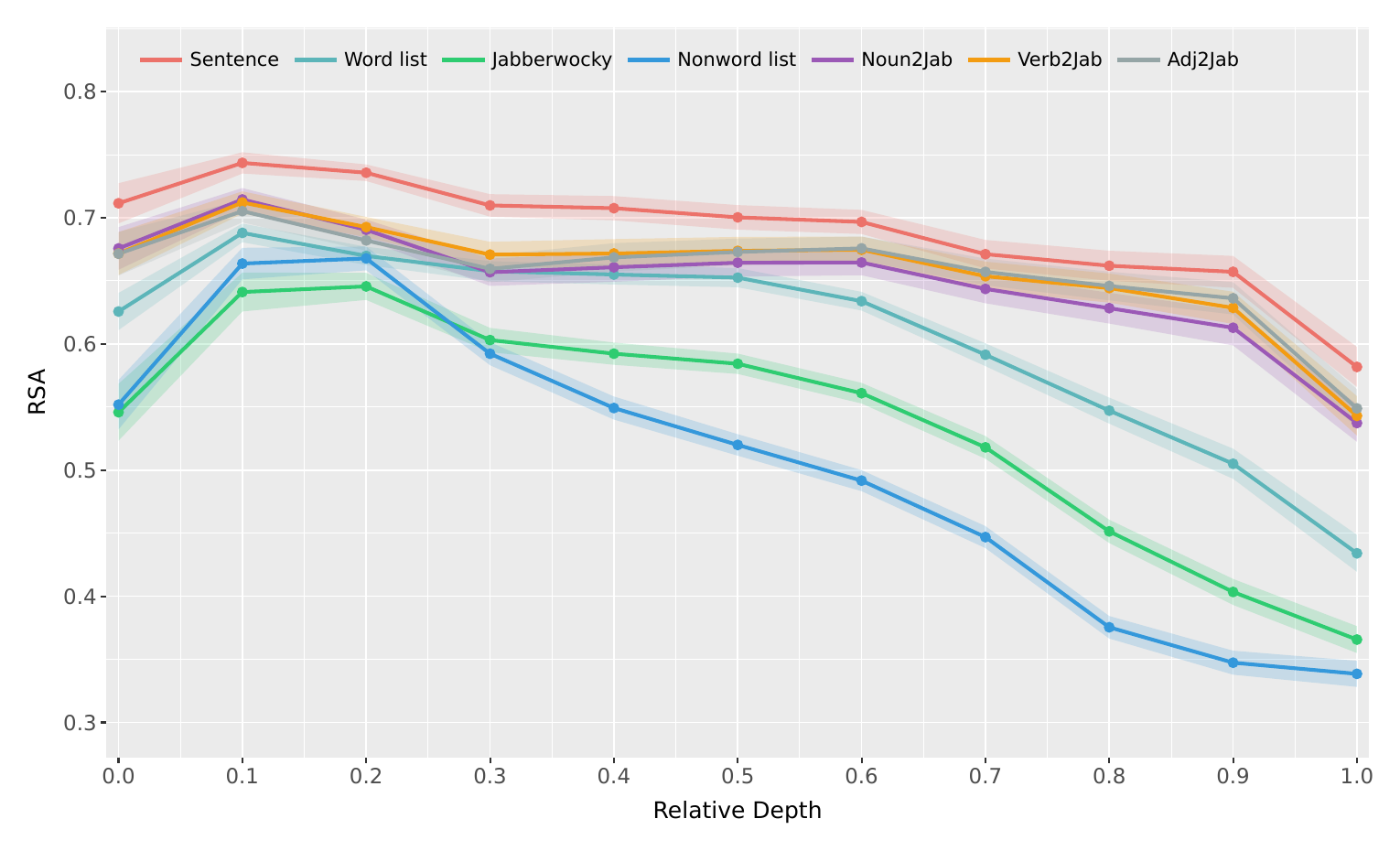}\hfill
\includegraphics[width=0.49\linewidth]{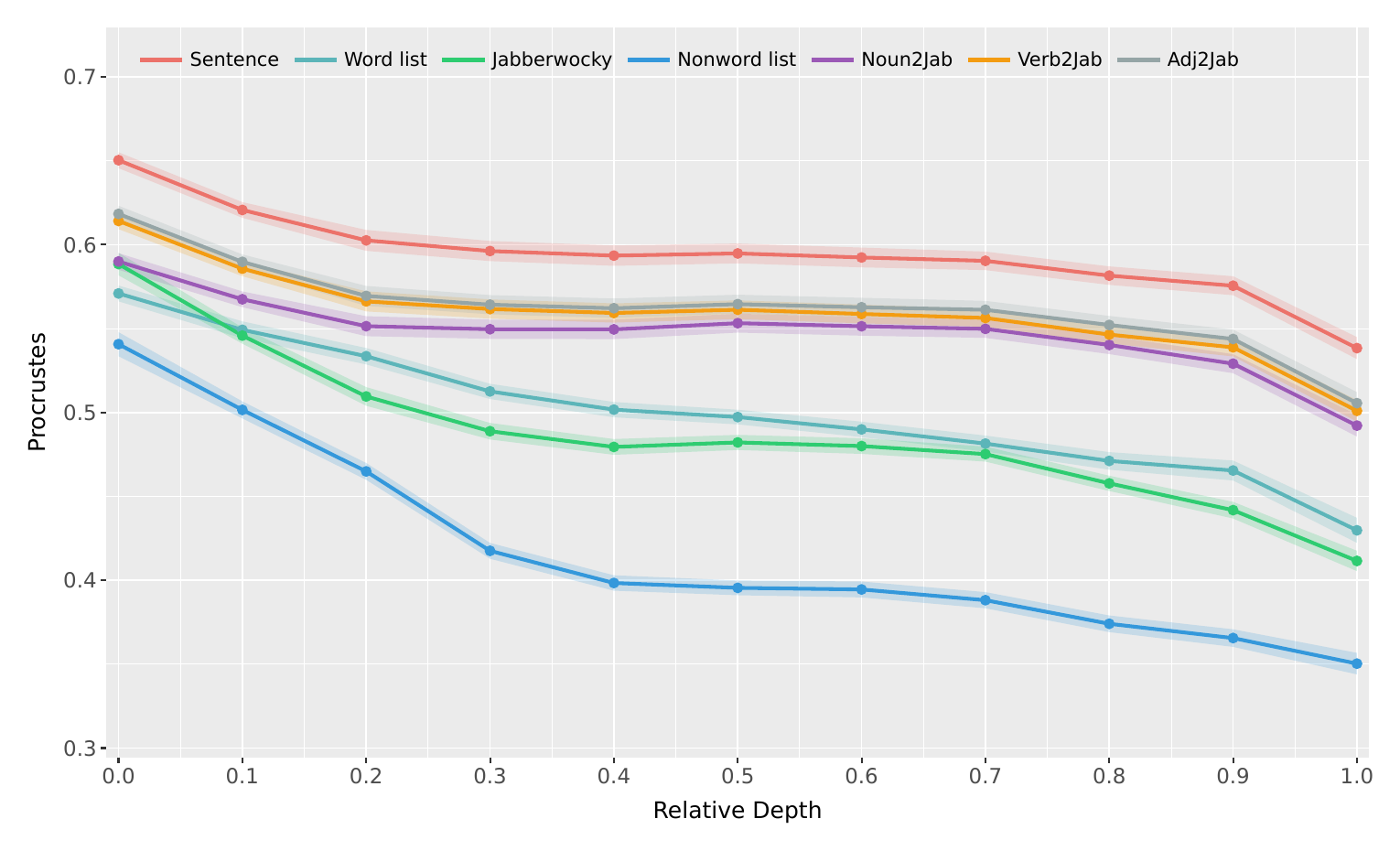}\\[4pt]
\includegraphics[width=0.49\linewidth]{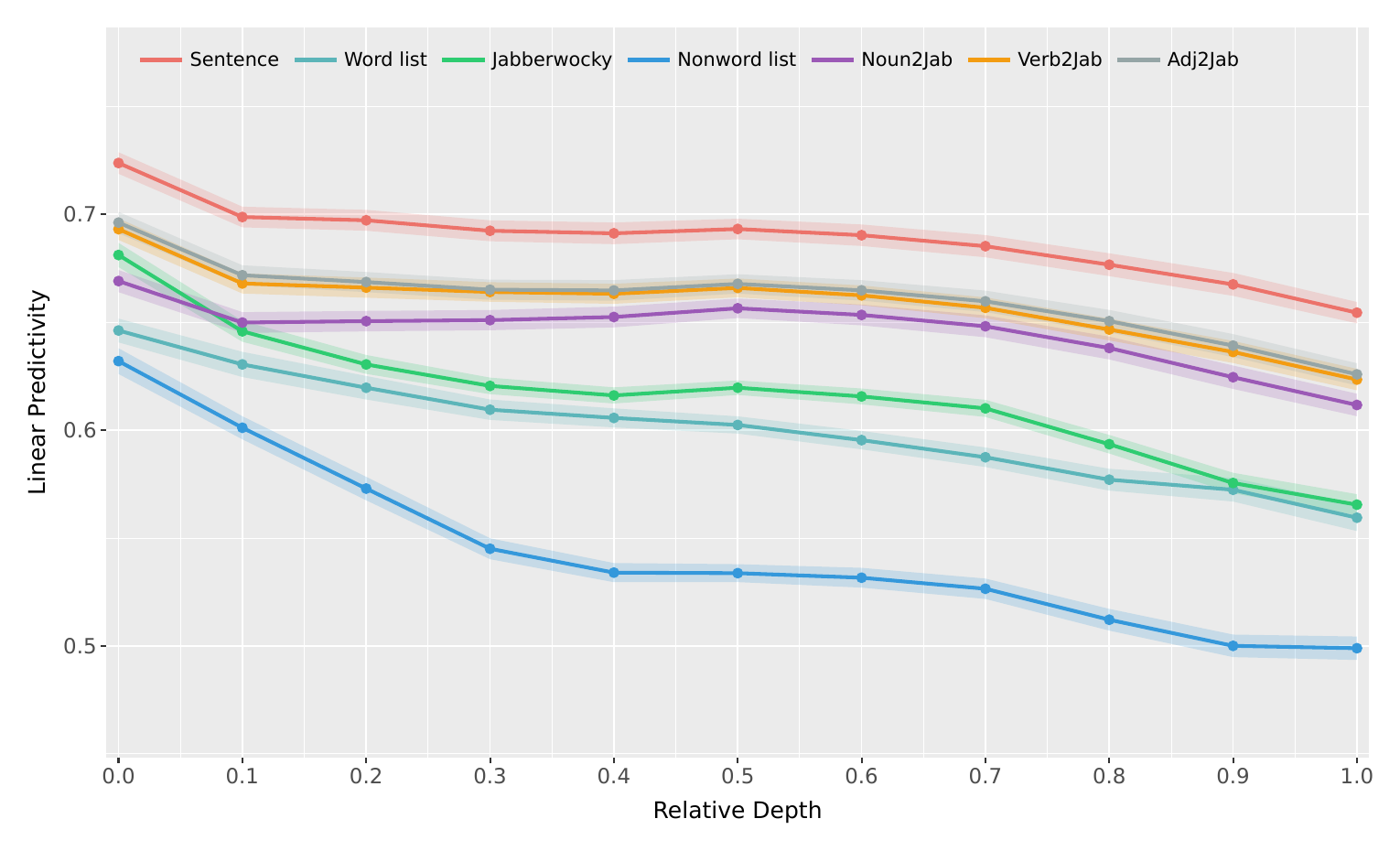}
\caption{Layer-wise off-manifold alignment by relative network depth. \textbf{Top left:} RSA; \textbf{top right:} Procrustes; \textbf{bottom:} linear predictivity. Natural sentences align best at every depth, and the gap to word lists, jabberwocky, and nonword lists is larger in later layers than in the first layer.}
\label{fig:offmanifold-layerwise}
\end{figure}

\begin{figure}[t]
\centering
\includegraphics[width=0.49\linewidth]{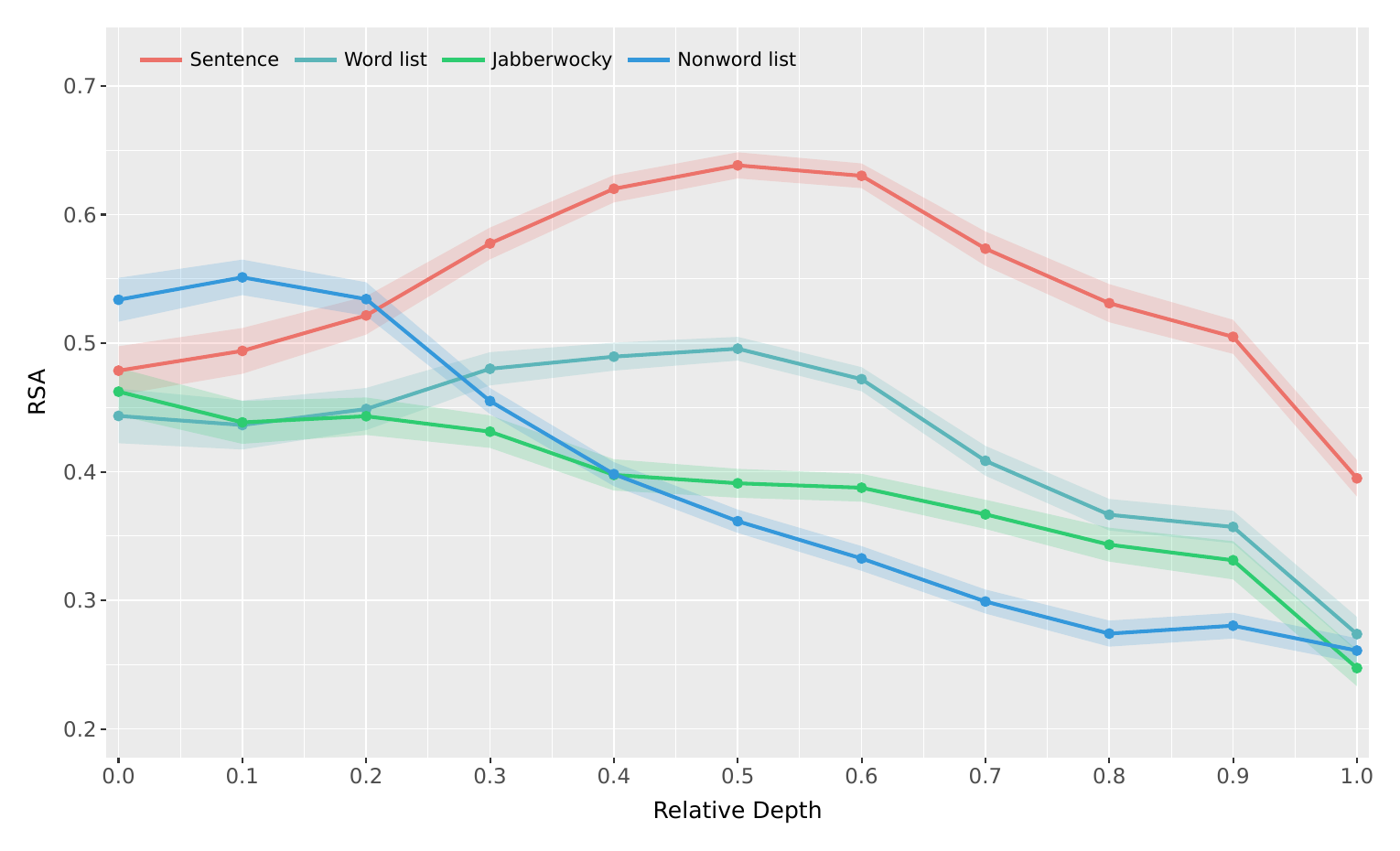}\hfill
\includegraphics[width=0.49\linewidth]{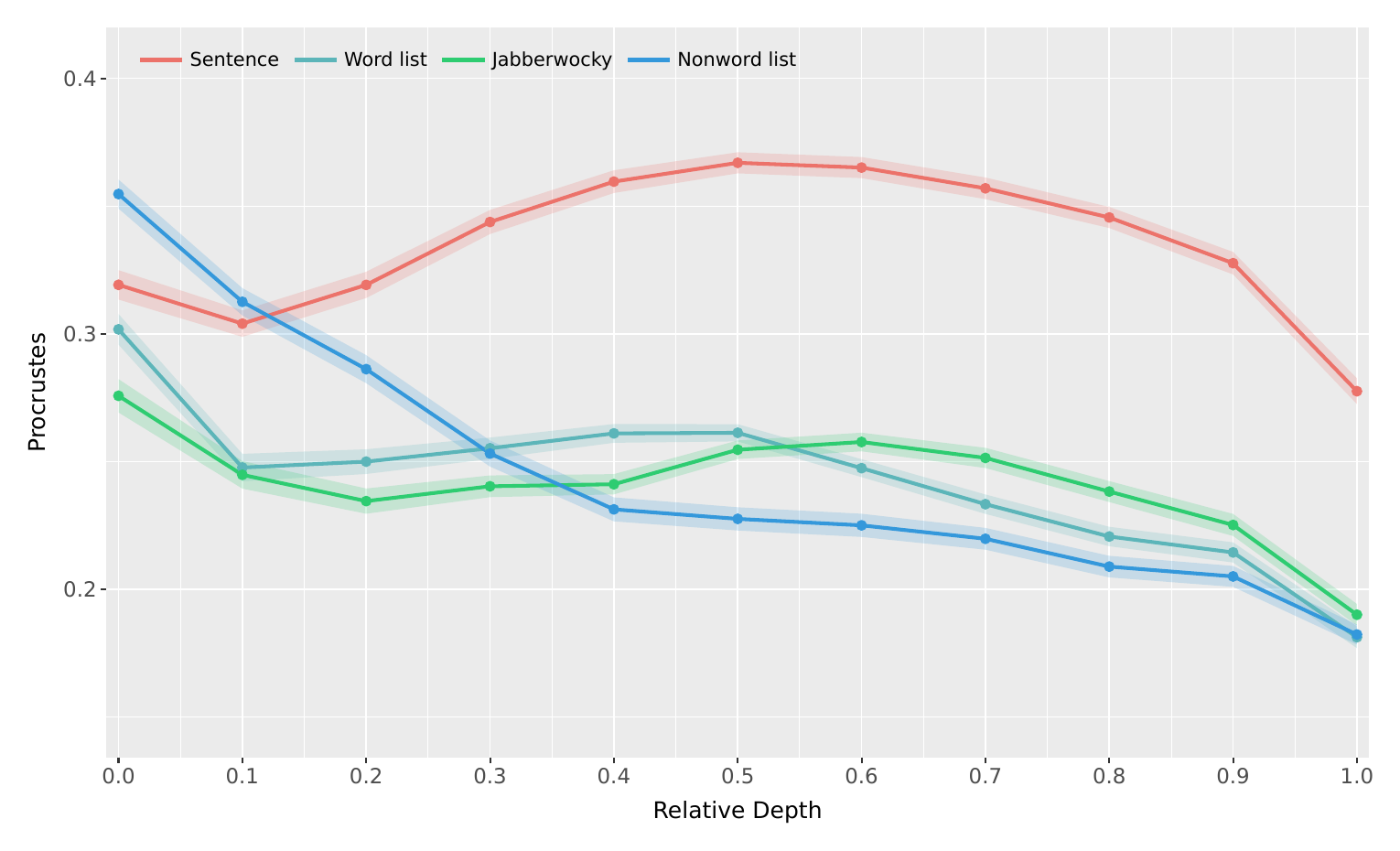}\\[4pt]
\includegraphics[width=0.49\linewidth]{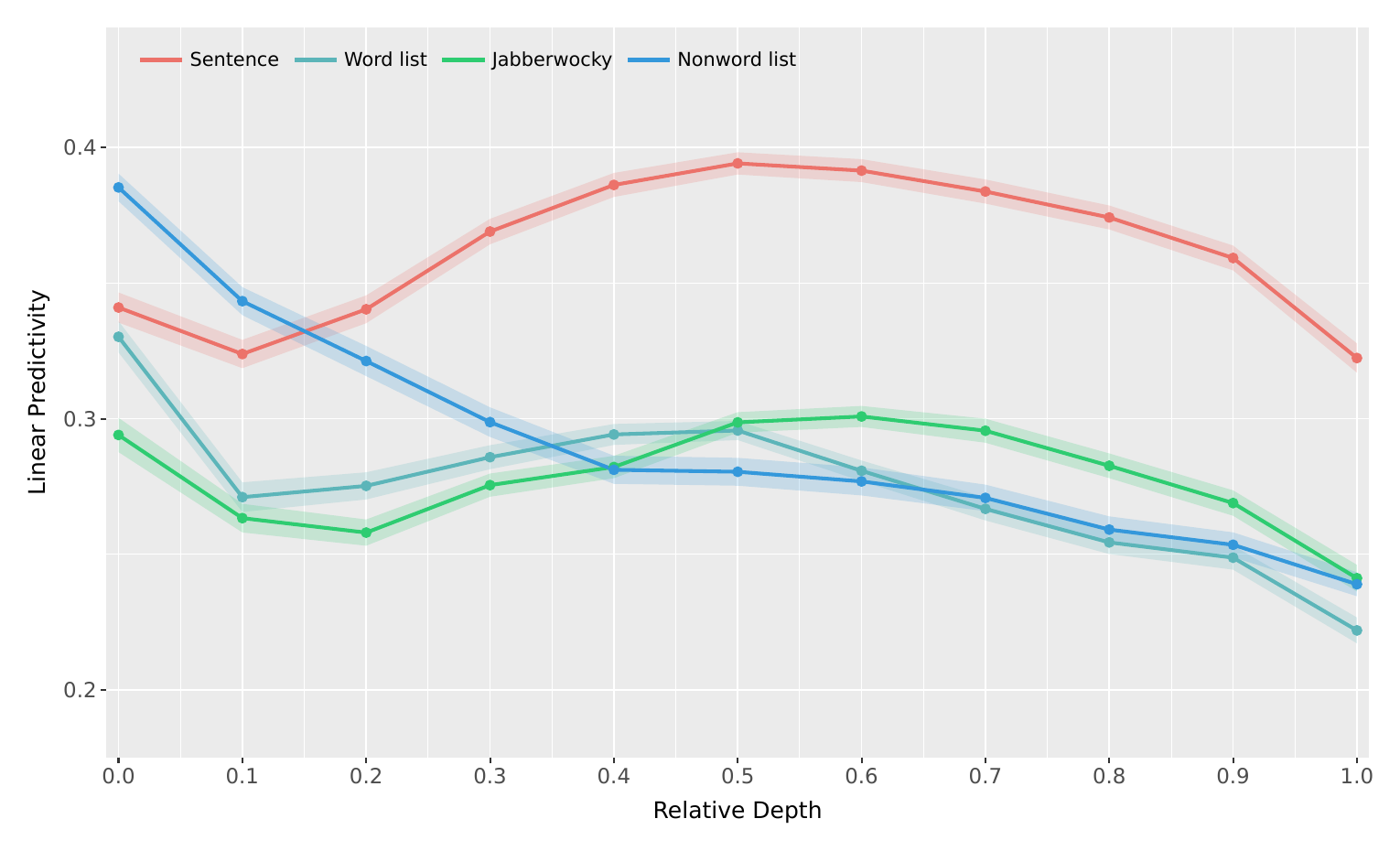}
\caption{Layer-wise off-manifold alignment on the human-authored stimuli of \citet{regev2024neural} by relative network depth. \textbf{Top left:} RSA; \textbf{top right:} Procrustes; \textbf{bottom:} linear predictivity. In the earliest layers, nonword lists align as well as or better than sentences; beyond them, sentences align best.}
\label{fig:swjn-layerwise}
\end{figure}

\paragraph{Reduced linguistic regularity.}
On our stimuli (Figure~\ref{fig:offmanifold-layerwise}), single-word replacements stay close to natural sentences at every depth. The gap between natural sentences and the conditions that remove structure or meaning changes differently across metrics. Under RSA, it narrows over the earliest layers and then widens until near the output; under Procrustes and linear predictivity, it increases from the first layer to the middle of the network and then stays roughly constant. On the human-authored stimuli of \citet{regev2024neural} (Figure~\ref{fig:swjn-layerwise}), sentence alignment peaks in the middle of the network rather than in the earliest layers.

\clearpage
\suppressfloats[t]
\section{Implementation Details}
\label{app:impl}

\paragraph{Representation extraction.}
For every model we run a single forward pass per stimulus and read hidden states from the penultimate layer (Section~\ref{sec:setup}). Appendix~\ref{app:layers} repeats the analyses at other depths. A sentence representation is the mean of hidden states over all non-special, non-padding tokens. All representations are mean-centered before any distance is computed, and we use correlation distance ($1-\text{Pearson}$) throughout.

\paragraph{Equivalence classes.}
For each equivalence class, RSA is computed separately among the $50$ neighbors of each of its $150$ groups and then averaged over groups; Figure~\ref{fig:linguistic-scale} (left) reports these averages. The global baseline is also computed within each class, over its $150$ anchor sentences (one per group); the Global bar in that panel is its average over the five classes.

\paragraph{NW-RSA and CF-RSA settings.}
RDMs use the upper-triangular pairs of the stimulus set. Neighborhood-Weighted RSA is evaluated across a range of localities: binary $k$-NN masks with neighborhood proportion $k_{\mathrm{prop}}\in\{0.005,0.01,0.02,0.05,0.1,0.2\}$ (equivalently $k$ from $50$ to $2000$ on the web corpora), and graded weights with temperature $\tau\in\{0.01,0.03,0.1,0.3,1.0\}$. Both recover global RSA as the locality knob opens to the full set. The geometric-scale sweep (Figure~\ref{fig:scale-results}) runs on two web corpora (DCLM, FineWeb, $10{,}000$ language-filtered sentences each) and the full stimulus set released by~\citet{tuckute2024driving} (\textsc{Control}). That study recorded fMRI responses of the human language network to $2{,}000$ short sentences, nearly all six words long. The set has three conditions. The $1{,}000$ baseline sentences are naturalistic sentences sampled to be diverse in topic and genre (e.g., news, web text, transcribed speech), and they were used to train a GPT-2-XL-based encoding model of the brain response. The $500$ drive and $500$ suppress sentences are those this encoding model predicts to evoke unusually high or low responses; in each condition, half were found by searching a large sentence corpus and half by iteratively editing sentences. Drive sentences are often ungrammatical or fragmentary (e.g., ``ASY would Azerbaijan subs taken MIT.''), whereas suppress sentences tend to be plain and frequent (e.g., ``You get out of the car.''). \textsc{Control}~(T) keeps only the $1{,}000$ baseline sentences (the original release labels this set T). It checks whether the local--global gap holds without the atypical drive and suppress items. Counterfactual RSA correlates model $A$'s intact RDM with model $B$'s within-model cross-condition RDM (Section~\ref{sec:counterfactual-rsa}). Each mismatch phenomenon uses $750$ matched item pairs, randomly sampled from the source-benchmark pairs that passed our screening (Appendix~\ref{app:safety}).

\paragraph{Statistics.}
Alignment is aggregated over model pairs and reported as the mean with $\pm1$ SEM over pairs. Matched-vs-counterfactual and on-vs-off-manifold contrasts are read from whether these intervals separate.

\paragraph{Compute.}
No model training or fine-tuning is performed: the pipeline runs forward passes for hidden-state extraction followed by lightweight matrix correlations. Experiments were run on an internal server with 48 AMD EPYC 9224 CPU cores and 768~GB RAM, using two NVIDIA A40 GPUs (48~GB each) with four CPU data-loading workers. The $7$B-parameter models are extracted in half precision, while encoder and sub-$3$B decoder models fit comfortably on a single GPU. RDM, NW-RSA, and CF-RSA computations run on CPU. Cached pretrained checkpoints, hidden states, and dissimilarity matrices used approximately $1.4$~TB of storage.

\paragraph{Software.}
Models are loaded with PyTorch and the Hugging Face Transformers library, and numerical analysis uses NumPy and SciPy. Linguistic preprocessing uses spaCy~\citep{honnibal2020spacy}: we run the transformer pipeline \texttt{en\_core\_web\_trf} (falling back to \texttt{en\_core\_web\_sm} when it is unavailable) with default settings, both for part-of-speech tagging, which selects the noun, verb, or adjective replaced in the single-POS off-manifold conditions, and for named-entity recognition flagging person mentions during content filtering (Appendix~\ref{app:safety}). Licenses are listed in Appendix~\ref{app:assets}.

\section{Existing Assets and Licenses}
\label{app:assets}

This work analyzes existing pretrained models with publicly available stimuli and standard open-source software. We use all assets for non-commercial academic research, credit the original creators, and do not redistribute third-party weights. Table~\ref{tab:assets} summarizes the main assets and their license or usage terms. For model families with per-size license differences (e.g., Qwen-2.5), the applicable terms follow each model card.

\paragraph{Consistency with intended use.} We verified, for every external artifact whose intended use is specified, that our use is consistent with it. The pretrained checkpoints are public research models. We use them only for non-commercial analysis of their hidden representations, with no fine-tuning, no weight modification, and no redistribution, which all of the licenses in Table~\ref{tab:assets} permit. The stimulus datasets were released by their creators for NLP research and benchmarking, and we use them solely for that purpose: STS-B, HANS, SICK, DiscoEval, and ROCStories for evaluation-style probing; PAWS, ROLE-1500, and the control set of~\citet{tuckute2024driving} as controlled stimulus sources; and the DCLM and FineWeb web corpora as naturalistic sentence pools. Several of these carry non-commercial or research-only terms (e.g., the control battery under CC BY-NC-ND 4.0, SICK under CC BY-NC-SA 3.0, ROCStories for research use). Our analysis-only, non-commercial use respects each of these.

\begin{table*}[t]
\caption{Main existing assets used in this work and their licenses or usage terms. Exact Hugging Face identifiers and versions will accompany the code released upon acceptance.}
\label{tab:assets}
\begin{center}
\small
\setlength{\tabcolsep}{6pt}
\renewcommand{\arraystretch}{1.15}
\begin{tabularx}{\linewidth}{@{}lXX@{}}
\toprule
Category & Item & License / Terms \\
\midrule
\textit{Models} & BERT, ALBERT, ELECTRA~\citep{devlin2019bert,lan2020albert,clark2020electra} & Apache-2.0 \\
& RoBERTa, MiniLM, DeBERTaV3~\citep{liu2019roberta,wang2020minilm,he2023debertav3} & MIT \\
& GPT-2~\citep{radford2019language} & MIT \\
& Mistral-7B~\citep{jiang2023mistral} & Apache-2.0 \\
& Qwen-2.5~\citep{yang2024qwen25} & Apache-2.0 (3B: Qwen Research License) \\
& Llama-3.2~\citep{dubey2024llama} & Llama 3.2 Community License \\
& Llama-2~\citep{touvron2023llama} & Llama 2 Community License \\
\midrule
\textit{Stimuli} & STS-B~\citep{cer2017semeval} & Mixed-source: CC BY-SA 3.0/4.0, MS Research corpus terms, EMM/EU legal notices, DEFT project terms \\
& PAWS~\citep{zhang2019paws} & Free use (Google), QQP subset\\
& Control sentences~\citep{tuckute2024driving} & CC BY-NC-ND 4.0 \\
& ROLE-1500~\citep{shivagunde-etal-2023-larger} & Research use (no license file) \\
& HANS~\citep{mccoy-etal-2019-right} & MIT \\
& SICK~\citep{marelli-etal-2014-sick} & CC BY-NC-SA 3.0 \\
& DiscoEval, ROCStories~\citep{chen-etal-2019-evaluation,mostafazadeh-etal-2016-corpus} & MIT (code), ROCStories research use \\
& DCLM-baseline~\citep{li2024datacomp} & CC BY 4.0 (research use only), Common Crawl ToU \\
& FineWeb~\citep{penedo2024fineweb} & ODC-By 1.0, Common Crawl ToU \\
\midrule
\textit{Software} & PyTorch & BSD-3-Clause \\
& Transformers & Apache-2.0 \\
& NumPy, SciPy & BSD-3-Clause \\
& spaCy~\citep{honnibal2020spacy}  & MIT \\
& Wuggy~\citep{keuleers2010wuggy} & MIT \\
& Llama Guard-4~\citep{inan2023llama,metallamaguard4} & Llama 4 Community License \\
\midrule
\textit{Graphics} & Twemoji emoji in Figure~\ref{fig:overview}~\citep{twitter_twemoji} & CC BY 4.0 \\
\bottomrule
\end{tabularx}
\end{center}
\end{table*}

\clearpage
\suppressfloats[t]
\section{Limitations}
\label{app:limitations}

Our model panel is limited to open-weight models for which hidden-state extraction is feasible, leaving generalization to closed and substantially larger models untested. The stimulus sets also have limitations. Web-derived datasets retain noise despite filtering, while generated stimuli may contain unintended variation or regularities introduced by the generating models. Separate generation and auditing models, together with manual checks of the equivalence-class data, reduce but do not eliminate these concerns (Appendix~\ref{app:stimuli}). Further, all stimuli are in English and cover a limited range of constructions and topics, so the results need replication with broader samples and other languages. The counterfactual results also concern specific perturbations, which may alter multiple input properties at once. They establish sensitivity to the tested changes but do not uniquely identify its cause.

\end{document}

%% file: math_commands.tex
\usepackage{amsmath,amsfonts,bm}

\def\eqref#1{equation~\ref{#1}}

\def\1{\bm{1}}

\DeclareMathAlphabet{\mathsfit}{\encodingdefault}{\sfdefault}{m}{sl}
\SetMathAlphabet{\mathsfit}{bold}{\encodingdefault}{\sfdefault}{bx}{n}



%% file: framework.tex
\subsection{Representational Comparison Methods}
\label{sec:metrics}

We propose Neighborhood-Weighted RSA for comparing alignment across geometric scales and Counterfactual RSA for testing which stimulus properties support agreement. Both methods apply generally to comparisons between neural representations; here, we use them to study language models. We additionally evaluate Procrustes and linear predictivity to test whether our conclusions hold under alternative definitions of alignment.

\paragraph{Representations and global RSA.}
\label{sec:rdm}
Let $\mathcal{X}=\{x_i\}_{i=1}^{N}$ denote a shared stimulus set, and let $f_A(x_i)\in\mathbb{R}^{d_A}$ and $f_B(x_i)\in\mathbb{R}^{d_B}$ denote selected-layer representations. Each model yields a representational dissimilarity matrix (RDM), $D^A_{ij}=d(f_A(x_i),f_A(x_j))$, and likewise for $B$, where $d$ is correlation distance (one minus the Pearson correlation) between sentence-level hidden states.

As a global reference, standard RSA~\citep{kriegeskorte2008representational} measures agreement between the RDMs using the Spearman correlation of their upper-triangular, off-diagonal entries, $\mathrm{RSA}_{\mathrm{global}}(A,B)=\rho_{\mathrm{S}}(\mathrm{vec}_{<}(D^A),\mathrm{vec}_{<}(D^B))$, where $\mathrm{vec}_{<}$ stacks entries with $i<j$. This summarizes agreement across all stimulus pairs but does not distinguish local neighborhoods from broader geometric organization.

\subsubsection{Neighborhood-Weighted RSA}
\label{sec:nwrsa}

We introduce Neighborhood-Weighted RSA (NW-RSA; Figure~\ref{fig:methods}, left) to compare local and global geometric alignment within a single framework. NW-RSA weights stimulus pairs according to their proximity in representational space. Varying these weights changes the geometric scale of comparison while retaining the same correlation measure, rather than switching between global RSA and a separate neighborhood-overlap metric.

Let $\mathcal{U}=\{(i,j):i<j\}$, and let $u_{ij}$ and $v_{ij}$ rank $D^A_{ij}$ and $D^B_{ij}$ over $\mathcal{U}$. We assign nonnegative weights $W_{ij}$, with $W=W^\top$ and $W_{ii}=0$. Define $\bar u_W=(\sum_{\mathcal{U}}W_{ij}u_{ij})/(\sum_{\mathcal{U}}W_{ij})$ and $\tilde u_{ij}=u_{ij}-\bar u_W$, and analogously for $v$. Then
\begin{equation}
\mathrm{NW\text{-}RSA}(A,B;W)=
\frac{\sum_{\mathcal{U}}W_{ij}\tilde u_{ij}\tilde v_{ij}}
{\sqrt{\left(\sum_{\mathcal{U}}W_{ij}\tilde u_{ij}^{2}\right)
\left(\sum_{\mathcal{U}}W_{ij}\tilde v_{ij}^{2}\right)}}.
\end{equation}
Ranks are computed once over all pairs and retained as the weights change. Thus, NW-RSA changes which pairs contribute without re-ranking the selected distances.

We use two weighting schemes. \emph{Binary neighborhood weights} retain a pair if either stimulus is among the other's $k$ nearest neighbors in either model:
\begin{equation}
W_{ij}=\mathbf{1}\!\left[
j\in\mathcal{N}^A_k(i)\lor j\in\mathcal{N}^B_k(i)
\lor i\in\mathcal{N}^A_k(j)\lor i\in\mathcal{N}^B_k(j)
\right],\qquad i\neq j.
\end{equation}
Each neighbor set excludes the stimulus itself, and increasing $k$ expands the retained pairs. \emph{Graded neighborhood weights} instead decrease continuously with distance, $W_{ij}=\sqrt{\exp(-D^A_{ij}/\tau)\exp(-D^B_{ij}/\tau)}$ for $i\neq j$. Smaller $\tau>0$ emphasizes nearby pairs, whereas larger values distribute weight more evenly.

At $k=N-1$, or as $\tau\to\infty$, all off-diagonal weights become one and NW-RSA reduces to global RSA. Alignment can increase, decrease, or vary nonmonotonically as neighborhoods expand, depending on agreement among newly emphasized pairs. Figure~\ref{fig:methods} illustrates both increasing and decreasing curves, so the empirical trend is not built into the definition.

\paragraph{Mapping-based alignment.}
RSA compares dissimilarity patterns without fitting a mapping between feature spaces. Procrustes~\citep{williams2021generalized} and linear predictivity~\citep{yamins2014performance} instead fit such a mapping and differ in the transformations they allow. Let $\mathbf{X}$ and $\mathbf{Y}$ contain intact-input representations from $A$ and $B$. We map $B$ to $A$, so that $B$ can subsequently receive perturbed inputs while $A$ remains the reference:
\begin{equation}
\widehat{\mathbf{M}}=\operatorname*{arg\,min}_{\mathbf{M}\in\mathcal{M}}
\lVert\mathbf{X}_{\mathrm{fit}}-\mathbf{Y}_{\mathrm{fit}}\mathbf{M}\rVert_{\mathrm{F}}^2.
\end{equation}
For equal-dimensional representations, Procrustes restricts $\mathcal{M}$ to orthogonal maps, comparing shape up to rotation and reflection. Linear predictivity permits unrestricted linear maps, including nonuniform scaling and shearing. Alignment is scored by the correlation between target representations and mapped predictions, $S_{\mathrm{intact}}=\operatorname{corr}(\mathbf{X}_{\mathrm{eval}},\mathbf{Y}_{\mathrm{eval}}\widehat{\mathbf{M}})$.

\subsubsection{Counterfactual Comparison}
\label{sec:counterfactual-rsa}

Alignment on intact inputs does not by itself reveal which stimulus properties contribute to agreement. We introduce Counterfactual RSA (CF-RSA; Figure~\ref{fig:methods}, right) to investigate this question through controlled input perturbations. In our linguistic experiments, the perturbations alter selected aspects of interpretation while preserving much of the surface lexical content.

\paragraph{Counterfactual RSA.}
Let $g$ denote a perturbation. CF-RSA compares model $A$'s intact RDM with a cross-condition dissimilarity matrix from model $B$, $D^{B,g}_{ij}=d(f_B(x_i),f_B(g(x_j)))$. Rows correspond to intact inputs and columns to their perturbed counterparts. The intact rows thus give a fixed reference for the perturbed representations.

Because $D^{B,g}$ is generally asymmetric, we compare all ordered off-diagonal entries:
\begin{equation}
\mathrm{CF\text{-}RSA}(A,B;g)=
\rho_{\mathrm{S}}\big(\mathrm{vec}_{\neq}(D^A),\mathrm{vec}_{\neq}(D^{B,g})\big),
\end{equation}
where $\mathrm{vec}_{\neq}$ stacks entries with $i\neq j$. We exclude the diagonal because it would compare zero self-distances in $A$ with intact and perturbed distances in $B$.

The decrease $\Delta_{\mathrm{CF}}=\mathrm{RSA}_{\mathrm{global}}(A,B)-\mathrm{CF\text{-}RSA}(A,B;g)$ quantifies perturbation sensitivity. When $g$ is the identity, CF-RSA equals global RSA. A larger decrease indicates greater sensitivity to the tested change, helping identify which properties support agreement.

\paragraph{Counterfactual evaluation of fitted mappings.}
To test whether the conclusions extend beyond RSA, we fit each mapping on intact inputs and keep it fixed during counterfactual evaluation. Let $\mathbf{Y}^{g}_{\mathrm{eval}}$ contain $B$'s representations of perturbed evaluation inputs. We compare $S_{\mathrm{perturbed}}=\operatorname{corr}(\mathbf{X}_{\mathrm{eval}},\mathbf{Y}^{g}_{\mathrm{eval}}\widehat{\mathbf{M}})$ with $S_{\mathrm{intact}}$. Model $A$ remains intact, and the mapping is not refitted. The change thus tests whether the correspondence learned on intact inputs survives the targeted change.

\subsection{Controlled Linguistic Stimulus Sets}
\label{sec:linguistic-locality}

Geometric neighborhoods group inputs by representational distance. To examine alignment within specific linguistic distinctions, we instead construct groups that hold selected linguistic properties fixed. These \emph{linguistic equivalence classes} define neighborhoods independently of whether their members are nearby in a model's representational space.

We evaluate five classes, each with $150$ groups of one anchor sentence and $50$ neighbors (Appendix~\ref{app:stimuli}).
\textbf{Semantic-equivalence} preserves propositional meaning while varying lexical choice and syntactic form.
\textbf{Syntactic-equivalence} preserves abstract syntactic form (clause type, constituent structure, grammatical functions, word order, valency) while varying content words and meaning.
\textbf{Entity-equivalence} preserves relational or schematic meaning while varying entity identities.
\textbf{Discourse-equivalence} holds discourse context fixed while varying sentences within that context.
\textbf{Task-equivalence} preserves the underlying task or correct response while varying phrasing or framing.

For each class, we average the RSA computed among the $50$ neighbors of each group and compare it with RSA over the same class's $150$ anchor sentences (Section~\ref{sec:result-linguistic-scale}). Holding the metric fixed while varying the stimulus set tests whether alignment extends beyond broad differences among heterogeneous inputs.